\documentclass[10pt,journal,compsoc]{IEEEtran}
\usepackage{amsmath,amsfonts}
\usepackage{algorithmic}
\usepackage{algorithm}
\usepackage{array}
\usepackage[caption=false,font=normalsize,labelfont=sf,textfont=sf]{subfig}
\usepackage{textcomp}
\usepackage{stfloats}
\usepackage{url}
\usepackage{verbatim}
\usepackage{graphicx}
\usepackage{multirow}
\usepackage{threeparttable}
\usepackage{bbding}
\usepackage{newfloat}
\usepackage{listings}
\usepackage{booktabs}
\ifCLASSOPTIONcompsoc
  \usepackage[nocompress]{cite}
\else
  \usepackage{cite}
\fi
\ifCLASSINFOpdf
\else
\fi
\begin{document}
%
\title{Cross-view Graph Contrastive Representation Learning on Partially Aligned Multi-view Data}
%
%
%
%
\author{Yiming~Wang,~Dongxia~Chang,~Zhiqiang~Fu,~Jie~Wen,~\IEEEmembership{Member,~IEEE,}
        and~Yao~Zhao,~\IEEEmembership{Senior~Member,~IEEE}
        
\IEEEcompsocitemizethanks{\IEEEcompsocthanksitem{Y. Wang, D. Chang, Z. Fu and Y. Zhao are with the Institute of Information
Science, Beijing Jiaotong University, Beijing 100044, China, and also
with Beijing Key Laboratory of Advanced Information Science and
Network Technology, Beijing 100044, China (e-mail: wangym@bjtu.edu.cn; dxchang@bjtu.edu.cn; zhiqiangfu@bjtu.edu.cn;yzhao@bjtu.edu.cn).}
\IEEEcompsocthanksitem{J. Wen is with the Shenzhen Key Laboratory of Visual Object Detection and Recognition, Harbin Institute of Technology, Shenzhen, Shenzhen China (jiewen\_pr@126.com)}
}
\thanks{(Corresponding author: Dongxia Chang)}}
\markboth{Journal of \LaTeX\ Class Files,~Vol.~14, No.~8, August~2015}%
{Shell \MakeLowercase{\textit{et al.}}: Bare Demo of IEEEtran.cls for Computer Society Journals}
%



\IEEEtitleabstractindextext{%
\begin{abstract}
Multi-view representation learning has developed rapidly over the past decades and has been applied in many fields. However, most previous works assumed that each view is complete and aligned. This leads to an inevitable deterioration in their performance when encountering practical problems such as missing or unaligned views. To address the challenge of representation learning on partially aligned multi-view data, we propose a new cross-view graph contrastive learning framework, which integrates multi-view information to align data and learn latent representations. Compared with current approaches, the proposed method has the following merits: (1) our model is an end-to-end framework that simultaneously performs view-specific representation learning via view-specific autoencoders and cluster-level data aligning by combining multi-view information with the cross-view graph contrastive learning; (2) it is easy to apply our model to explore information from three or more modalities/sources as the cross-view graph contrastive learning is devised. Extensive experiments conducted on several real datasets demonstrate the effectiveness of the proposed method on the clustering and classification tasks.
\end{abstract}

}

\maketitle

\IEEEdisplaynontitleabstractindextext

%
\IEEEpeerreviewmaketitle

\IEEEraisesectionheading{\section{Introduction}\label{sec:introduction}}

%
%
%
%
\IEEEPARstart{I}{n} real-world applications, data often appears in multiple modalities or comes from multiple sources. Due to the variability in data acquisition, there is a highly heterogeneous difference between the same instance in different views \cite{conf/colt/BlumM98, conf/cvpr/ZhangLF19}. Simultaneously, the characteristics of the different views provide complementary and consistent information \cite{9258396}. For example, a sporting event can be reported by various mediums, such as text, images, and video. The text can come from different media organizations or even different languages. The images can also be described with various color and texture descriptors. Consequently, many methods have been proposed to explore the comprehensive information of multi-view data to boost the performance of multi-view learning.

Current multi-view learning methods generally extract helpful information from multiple views to learn common metrics or representations for downstream tasks. According to the prior difference, it can be divided into supervised multi-view learning and unsupervised multi-view representation learning. A representative supervised multi-view learning example is multi-view metric learning \cite{conf/ijcai/ZhangLLHLZ18}, which jointly learns a set of distance metrics for multi-view data. In addition, as the representative techniques of unsupervised multi-view representation learning, Canonical Correlation Analysis (CCA) \cite{cca} computes shared embeddings of multi-view data by maximizing the correlation between them. Due to its capability of effectively modeling the relationship between different views, numerous extended CCA approaches have been proposed over the past few decades, including Sparse CCA \cite{conf/icml/dAspremontBG07}, Kernel CCA \cite{journals/ijns/LaiF00}, and Deep CCA \cite{conf/cvpr/YanM15}. Particularly, Deep CCA methods use deep neural networks to learn the representation of multiple views by maximizing inter-view relations.

\begin{figure}[t]
    \centering
    \subfloat[]{\includegraphics[width=2cm]{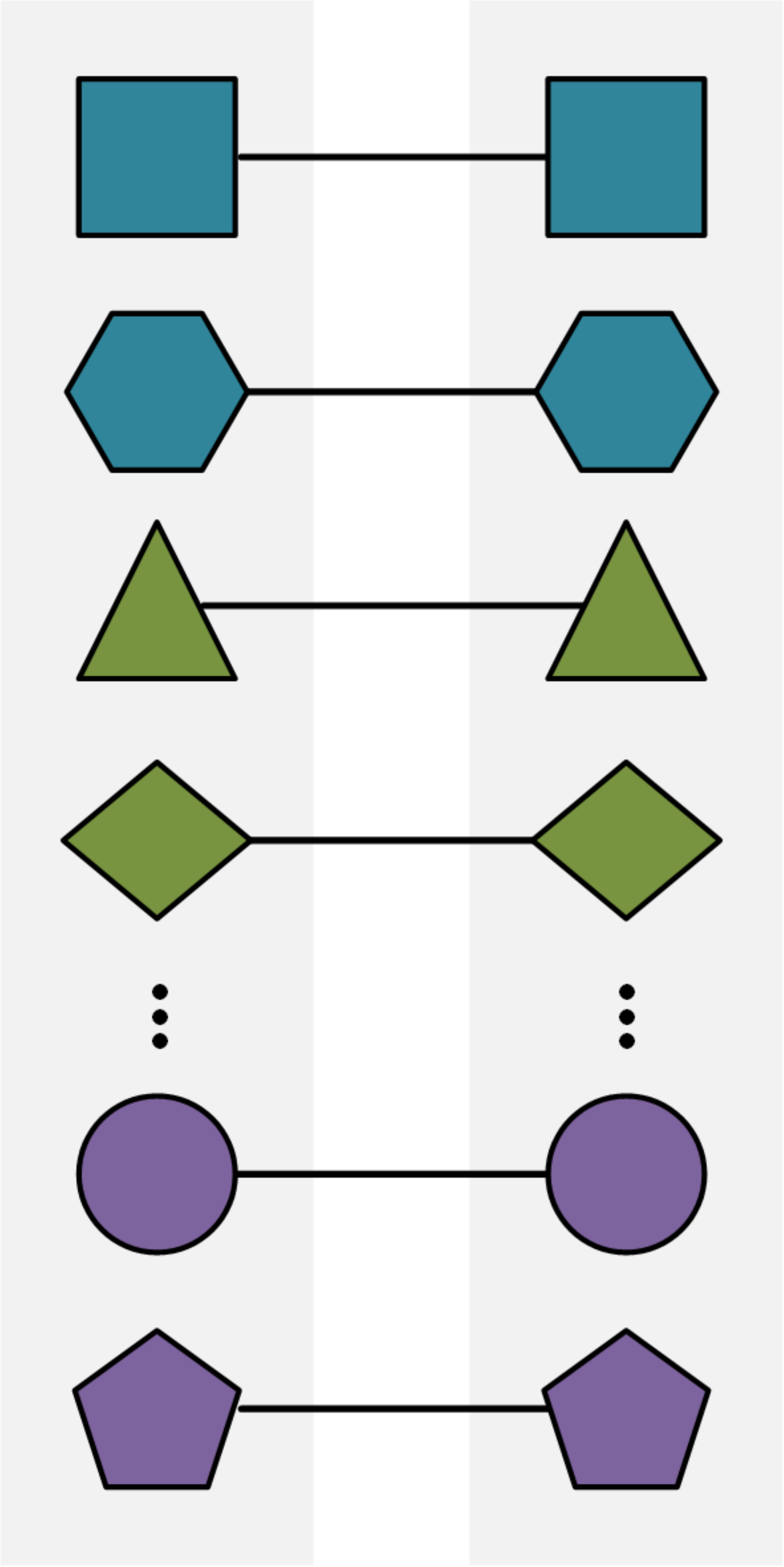}\label{fig:align}}\hspace{1mm}
    \subfloat[]{\includegraphics[width=2cm]{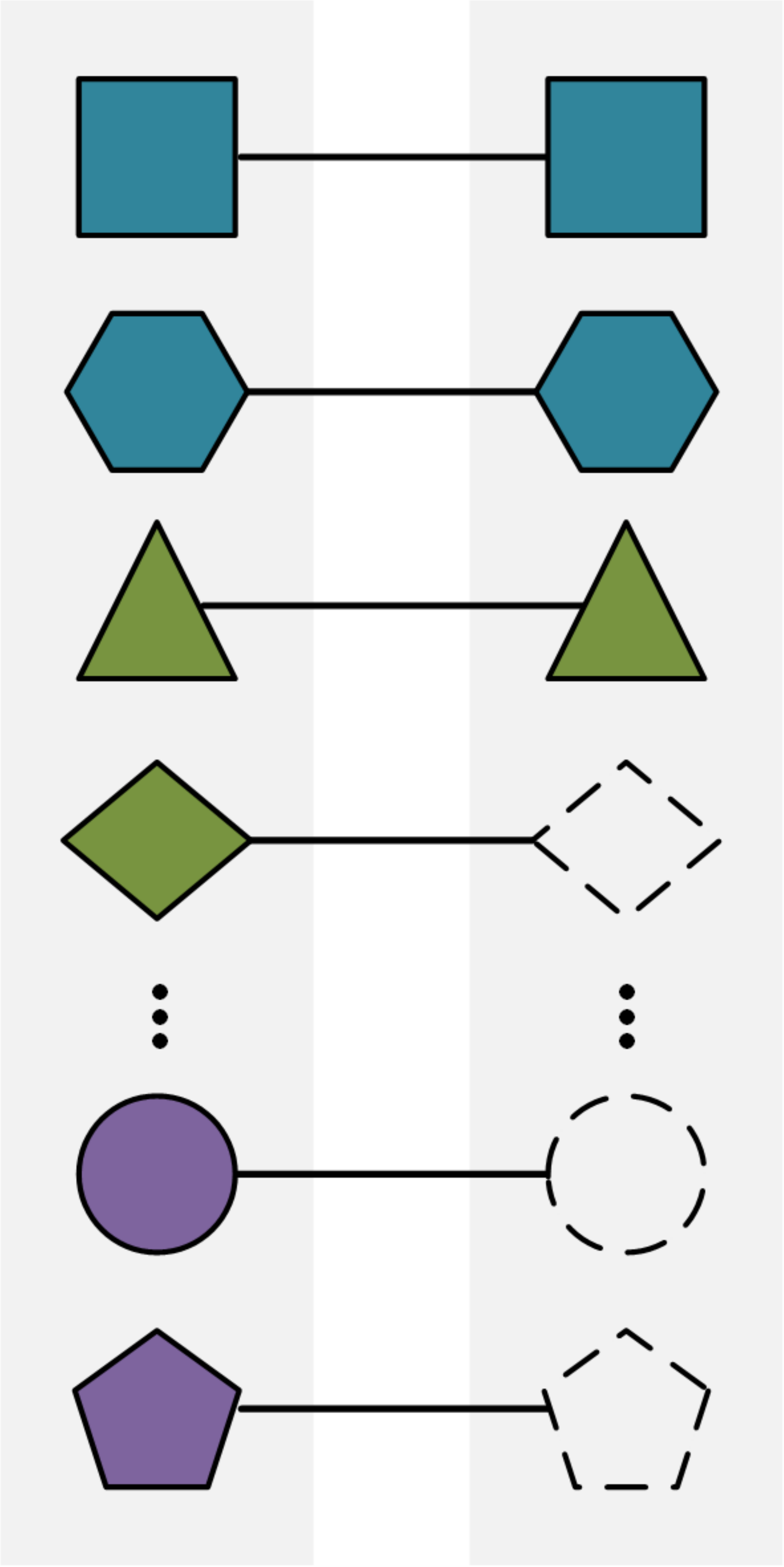}\label{fig:incomplete}}\hspace{1mm}
    \subfloat[]{\includegraphics[width=2cm]{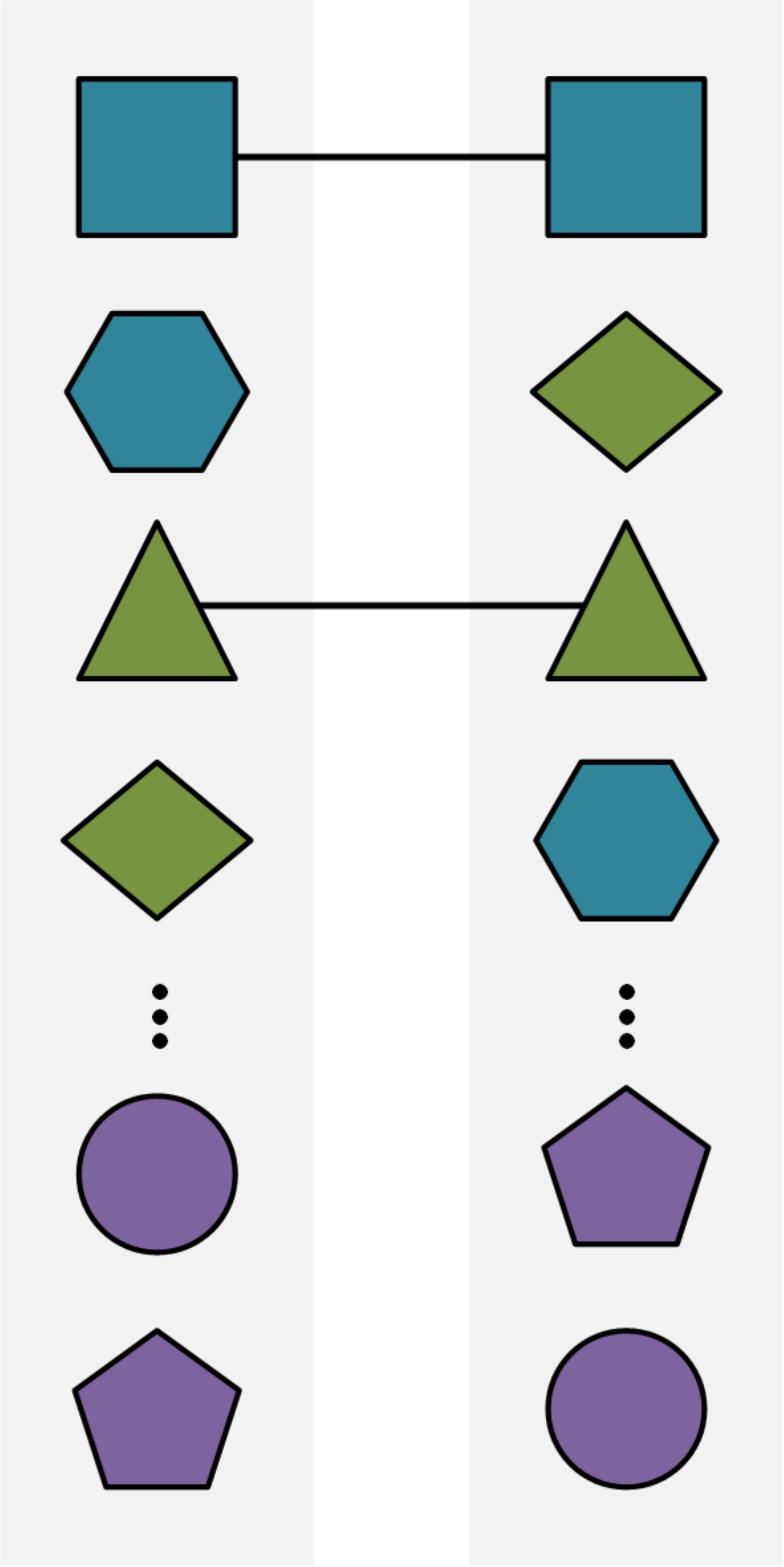}\label{fig:unaligned}}\hspace{1mm}
    \subfloat[]{\includegraphics[width=2cm]{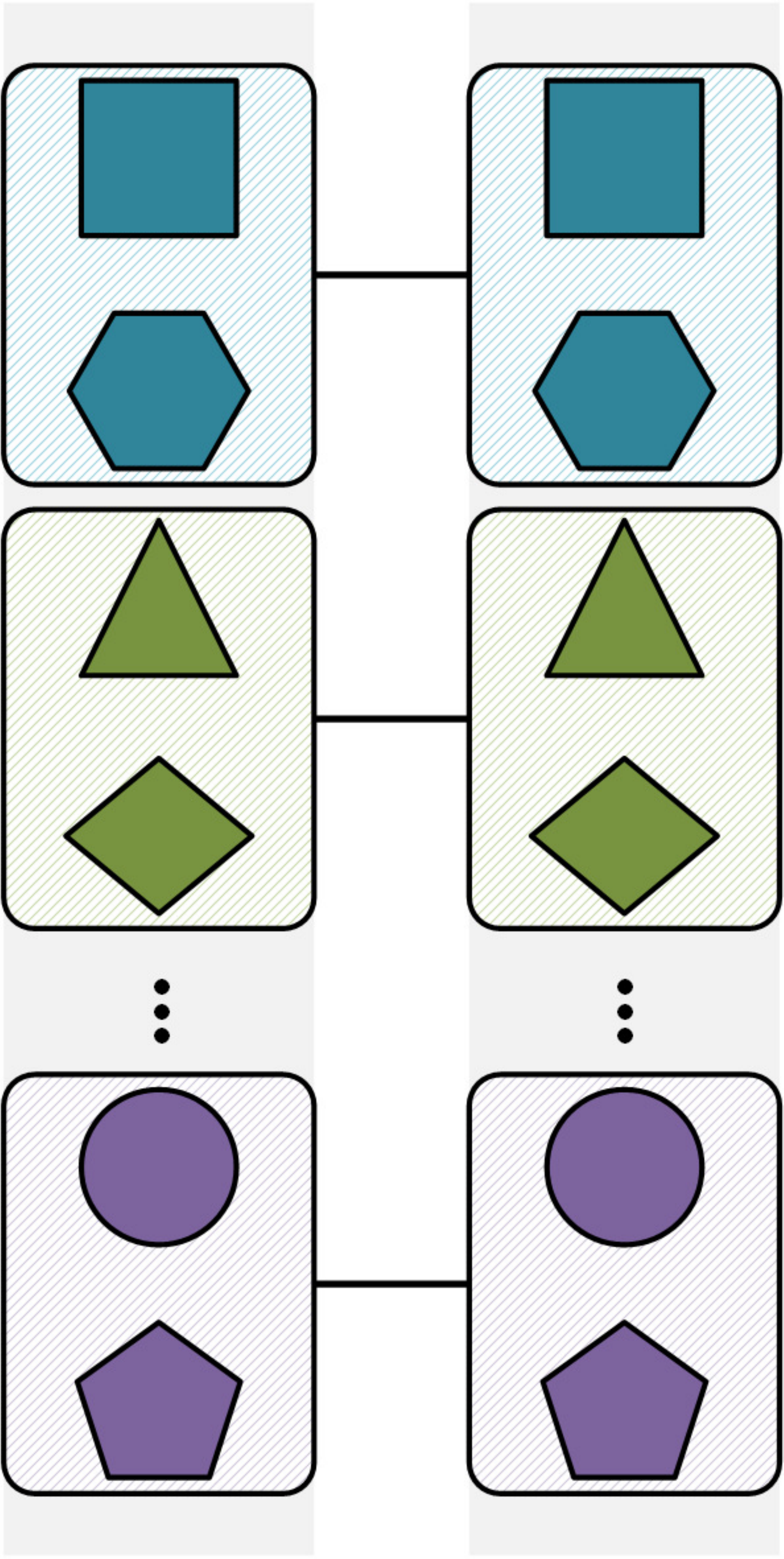}}\hspace{1mm}
    \caption{Various multi-view data. In the figure, different shapes denote different instances, different colors indicate different categories, and the line denotes the correspondence between multi-view data. (a) Multi-view data with complete and aligned views; (b) Incomplete multi-view data; (c) Partially aligned multi-view data; (d) Multi-view data with cluster-level alignment. Compared with instance-level alignment, cluster-level alignment is more easily achieved and suitable for downstream tasks such as clustering and classification. In this paper, cross-view graph contrastive learning is devised to perform cluster-level alignment and representation learning simultaneously.}
\end{figure}

Although current multi-view representation learning methods achieve promising performance, a common assumption for most methods is that all of the views are complete and aligned (see Fig.~\ref{fig:align}). However, in practical applications, multi-view data may be partially lost during transmission and storage, resulting in incomplete multi-view data (see Fig.~\ref{fig:incomplete}) \cite{conf/mm/0001ZZWFXZ20, conf/cvpr/0001GLL0021}. Additionally, spatial, temporal, or spatiotemporal asynchronism can cause some data to be unaligned across views, resulting in partially aligned multi-view data (see Fig.~\ref{fig:unaligned}) \cite{conf/nips/Huang0ZL020, conf/cvpr/YangL0L0021}. These conditions can lead to inevitable degeneration for most multi-view representation learning methods. To solve the problem of view incomplete, plenty of approaches have been proposed to reduce the impact of missing views using Non-negative Matrix Factorization (NMF) \cite{aaai/LiJZ14, conf/aaai/HuC19} or Generative Adversarial Networks (GANs) \cite{journals/corr/GoodfellowPMXWOCB14, 9258396}. However, few methods are proposed to solve the misalignment problem \cite{conf/nips/Huang0ZL020,conf/cvpr/YangL0L0021}. Representatively, Partially View-aligned Clustering (PVC) \cite{conf/nips/Huang0ZL020} proposes a differentiable Hungarian network to establish the correspondence of two views. However, the instance-level alignment implemented by \cite{conf/nips/Huang0ZL020} is hard to achieve and is over-sufficient for downstream tasks such as clustering and classification. Yang et al. \cite{conf/cvpr/YangL0L0021} further propose a step-wise alignment model that uses aligned samples as training samples to learn view correspondences via a noise-robust contrastive loss. These two approaches achieve the desired performance but can only be applied to two views due to their aligning processes.


In this paper, we propose a new \textbf{C}ross-v\textbf{I}ew g\textbf{R}aph \textbf{C}ontrastive representation \textbf{LE}arning (CIRCLE) framework to address problems in partially aligned multi-view learning. Unlike existing methods, CIRCLE is an end-to-end model that enables both cluster-level alignment and representation learning based on intra-cluster and inter-view consistency. Specifically, we construct relation graphs based on the distance between each sample and the aligned samples in the same view and maximize the similarity between the representation of the sample and its aligned neighboring samples in each view. The main contributions of the proposed CIRCLE are summarized as follows:
\begin{itemize}
    \item  We propose a novel end-to-end partially aligned multi-view representation learning model. As far as we know, CIRCLE is the first deep network that can handle partially aligned multi-view data with more than two views. 
    \item To perform cluster-level aligning and learn latent representation simultaneously, we devise a cross-view graph contrastive learning module based on intra-cluster and inter-view consistency.
    \item We have conducted clustering and classification experiments to evaluate the performance of CIRCLE. Compared with the existing state-of-the-art methods, CIRCLE achieves considerable improvement, demonstrating its effectiveness on partially aligned multi-view data.
\end{itemize}

The rest of the paper is organized as follows. Section~\ref{SECRelatedWork} briefly reviews the related work about contrastive learning and multi-view representation learning. In Section~\ref{SECProposed}, we present the proposed multi-view representation learning model, named CIRCLE. In Section~\ref{SECExperiments}, we validate the effectiveness of the proposed method with extensive experiments. Finally, we conclude this paper in Section~\ref{SECConclusion}.

\section{Related Works}
\label{SECRelatedWork}
Before introducing the proposed CIRCLE, we briefly review the related contrastive learning and unsupervised multi-view representation learning.

\subsection{Contrastive Learning}

Contrastive learning has become one of the most popular research topics in unsupervised learning due to its outstanding performance in many tasks \cite{9462394, conf/nips/Tian0PKSI20, 9667296}. The basic idea of contrastive learning is to learn a mapping that can map the raw data to a latent space in which the similarity of positive pairs is maximized and the similarity of negative pairs is minimized \cite{conf/cvpr/HadsellCL06}. One of the most commonly used pairs construction strategies is that the positive pair is composed of two augmentations of the same instance, and the other pairs are regarded as negative pairs. Next, to maximize the similarity between positive pairs and minimize the similarity between negative pairs, some contrastive losses are proposed  \cite{journals/jmlr/GutmannH10,conf/cvpr/He0WXG20,conf/icml/ChenK0H20}. There are also some methods to construct pairs through other strategies. For example, Sharma et al. \cite{conf/fgr/0001TSS20} propose a cluster-based contrastive learning approach, which constructs positive and negative pairs of samples using the cluster labels and trains a Siamese network to learn the face representation. \cite{conf/iccv/ZhongW0HDNL021} proposes a graph Laplacian-based contrastive loss and a graph-based contrastive learning strategy to maximize the similarity of representations and clustering assignments for samples in one cluster and their augmentations. Different from existing contrastive learning approaches, we focus on maximizing the similarity of the representations of similar instances from different views. Thus, we construct pairs using characteristics of similar instances in different views instead of data augmentation. By maximizing the similarity between positive pairs and minimizing the similarity between negative pairs, our approach can achieve data alignment and discriminative representation learning.

\subsection{Unsupervised Multi-view Representation Learning}

Due to the difficulty of obtaining multi-view data labels in practical applications, unsupervised multi-view representation learning has attracted increasing attention in the last few years \cite{8471216}. In early studies of multi-view representation learning, CCA \cite{cca} and its extensions like Sparse CCA \cite{conf/icml/dAspremontBG07} and Kernel CCA \cite{journals/ijns/LaiF00} are extensively studied \cite{8493362,7123622}. With the development of Deep Neural Networks (DNNs), CCA-like objectives are applied to neural networks to capture relations from multi-view data \cite{conf/icml/AndrewABL13, conf/cvpr/YanM15}. Further, Wang et al. \cite{conf/icml/WangALB15} combine Deep CCA and autoencoders (AEs) and optimizes the canonical correlation between the learned representations and the reconstruction errors simultaneously. The subsequent multi-view representation learning approaches explore multi-source information by presenting various complex network models and specific objective functions. For example, Autoencoder in Autoencoder Networks (AE$^2$-Nets) \cite{conf/cvpr/ZhangLF19} jointly performs view-specific representation learning and multi-view encoding by the nested autoencoder framework. With the development of multi-view representation learning techniques, more and more approaches are focusing on practical issues such as missing and unaligned data. To handle incomplete multi-view data, Zhang et al. \cite{9258396} propose a framework to simultaneously impute missing views and learn common representations of multiple views through an adversarial strategy. In addition, to handle the partially view-aligned problem, \cite{conf/cvpr/YangL0L0021} propose a noise-robust contrastive loss to reduce the impact of false negative pair on aligning data and representation learning.

\section{The Proposed Method}
\label{SECProposed}

\begin{figure*}[!t]
    \centering
    \includegraphics[width=\textwidth]{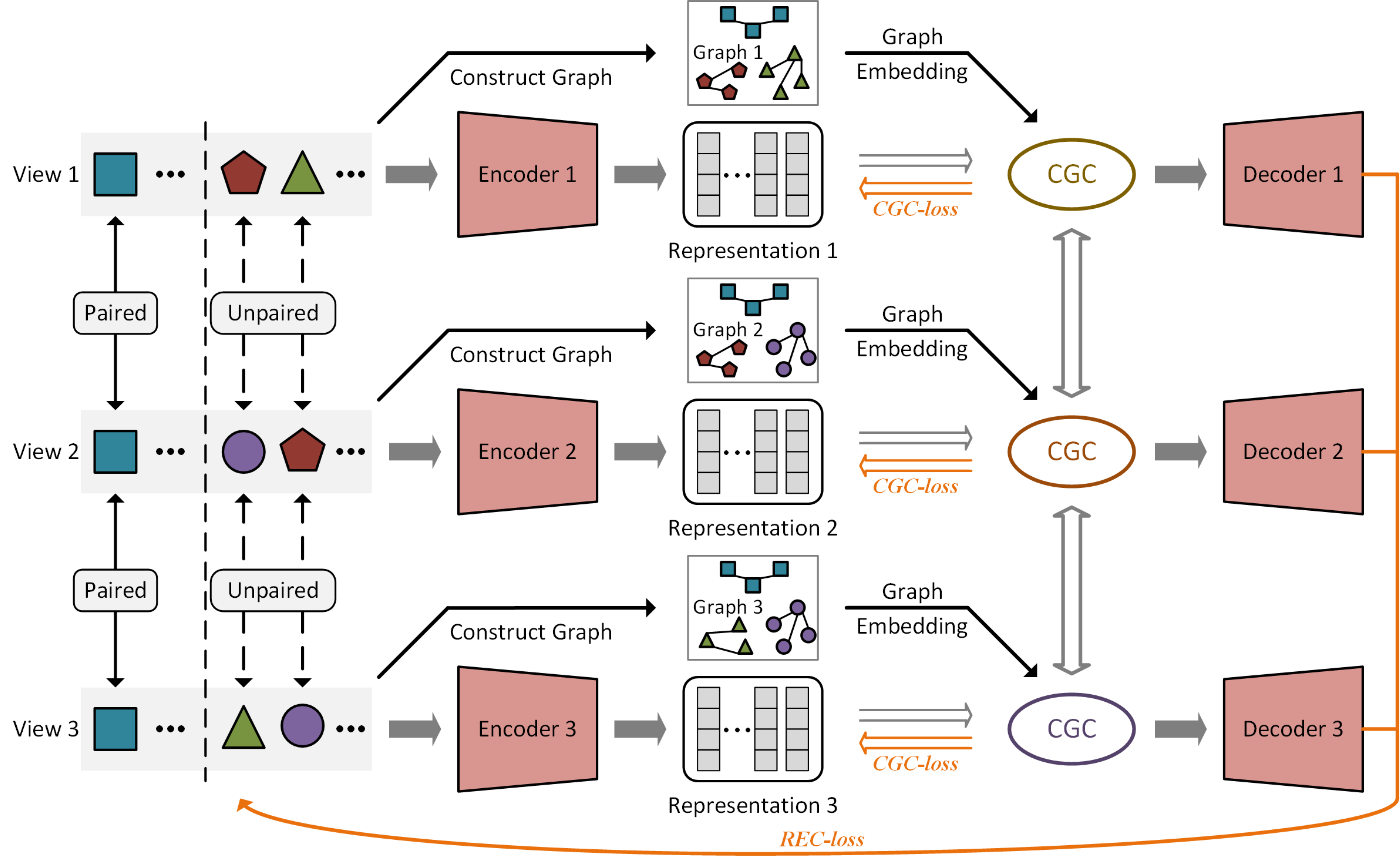}
    \caption{The framework of the proposed CIRCLE. In the figure, three-view data is used as a showcase. As shown, the cross-view graph contrastive loss and view-specific reconstruction loss are introduced to optimize the representations learned by a set of view-specific autoencoders.}
    \label{fig:model}
\end{figure*}

\subsection{Motivation and Notations}
As aforementioned analyses, there are still some problems to be solved in handling partially aligned multi-view data. On the one hand, due to the non-alignment of views, it is difficult to establish correlations between different views to learn consistent representations directly. On the other hand, it is challenging to extend the alignment operation to data with three or more modalities/sources, and current methods can only be applied to two views. To address these issues, we propose a new partially aligned multi-view representation learning method based on cross-view graph contrastive learning, termed CIRCLE. As illustrated in Fig.~\ref{fig:model}, the proposed CIRCLE consists of two main modules, i.e., view-specific autoencoders and cross-view graph contrastive learning module (CGC).

In the partially aligned multi-view learning, only a portion of instances is aligned across different views. Formally, given a partially aligned multi-view dataset $\{X^{(1)},...,X^{(V)}\}$ consisting of $N$ instances of $V$ views. $X^{(v)} = \{x^{(v)}_1,...,x^{(v)}_N\} \in \mathbb{R}^{ {d_v} \times N}$ denotes the feature matrix in the $v$-th view, where $d_v$ is the feature dimension of samples in the $v$-th view. For convenience, matrix $A$ is used to record the aligned instances, where $A_{i} = 1$ means the $i$-th instance is aligned in all views, otherwise $A_{i} = 0$.

\subsection{Backbone of CIRCLE: View-specific Autoencoders}

As shown in Fig.~\ref{fig:model}, the basic structure of CIRCLE is a set of view-specific autoencoders. There are several reasons to use view-specific autoencoders as the backbone. Firstly, the data from different views generally have varying dimensionality and contain distinctive characteristics that other views do not have. Secondly, it is hard to process data from multiply views together as the views are not aligned. To this end, view-specific autoencoders are applied to capture the view-specific features of different views. Following the previous multi-view learning approaches \cite{conf/ijcai/WenZ0ZFX20}, we set the encoder and decoder of each view as four stacked fully connected layers with ReLU activation function.

Let $E^{(v)}(x^{(v)}_n;\theta^{(v)}_E)$ and $D^{(v)}(z^{(v)}_n;\theta^{(v)}_D)$ denote view-specific encoders and decoders, where $\theta^{(v)}_E$ and $\theta^{(v)}_D$ are the network parameters, respectively. The learned latent representations of $v$-th view can be obtained by
\begin{equation}
z^{(v)}_n = E^{(v)}(x^{(v)}_n) \in \mathbb{R}^{d_{z}}
\end{equation}
For the subsequent cross-view graph contrastive learning, the representation of each view is set to the same dimension $d_{z}$. Then,
the view-specific decoders reconstruct the sample as $\hat{x}^{(v)}_n$ by decoding $z^{(v)}_n$:
\begin{equation}
\hat{x}^{(v)}_n = D^{(v)}(z^{(v)}_n) = D^{(v)}(E^{(v)}(x^{(v)}_n)) \in \mathbb{R}^{d_v}
\end{equation}
For the $v$-th view, the reconstruction loss $L_r^{(v)}$ between input $x^{(v)}_n$ and output $\hat{x}^{(v)}_n$ is optimized to map the raw data into the latent space. Then the total reconstruction loss can be written as

\begin{equation}
L_{REC} = \sum_{v=1}^V L_r^{(v)} = \sum_{v=1}^V\sum_{n=1}^N ||\hat{x}_n^{(v)}- x^{(v)}_n||^2_F
\end{equation}

\subsection{Cross-view Graph Contrastive Learning}
\begin{figure*}[!t]
    \centering
    \includegraphics[width=\textwidth]{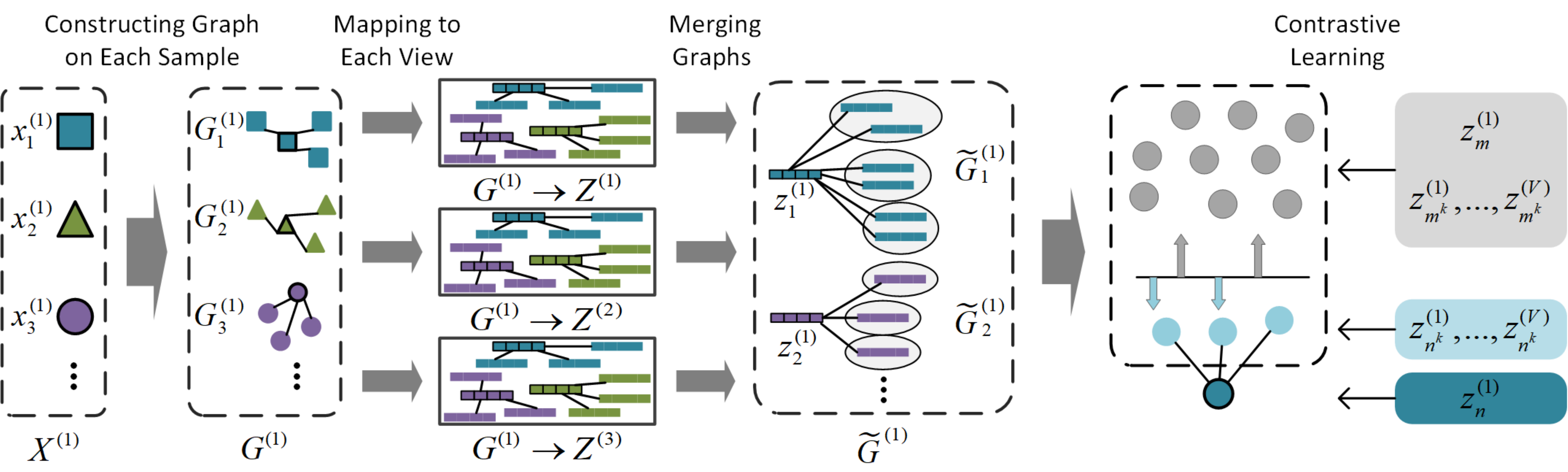}
    \caption{The pipeline of the cross-view graph contrastive learning module. The first of three views is used as a showcase in the figure. We first construct relation graphs centered on each sample. Then the aligned neighboring samples are mapped to other views based on the alignment relationship. Finally, we use a distance-weighted graph contrastive loss to train the model to maximize the similarity of representations of $x_n^{(v)}$ and samples in  $\Tilde{G}^{(v)}_n$.}
    \label{fig:cgc}
\end{figure*}

In order to simultaneously learn latent representations and acquire cluster-level alignment, we design a cross-view graph contrastive learning module (CGC) to lift the commonly-used instance-level contrastive learning strategies to cluster-level contrastive learning on multi-view data. 

Unlike previous contrastive learning approaches, we construct positive and negative pairs according to pseudo-labels generated by the relation graphs, which are constructed based on the distance of the samples in the original space. Based on the assumption that close samples in the original space have a high probability of belonging to the same category, we consider that if one sample is in the relation graph of another sample, they can form positive pair sharing the same pseudo-label, otherwise the opposite.

Fig.~\ref{fig:cgc} shows the pipeline of our cross-view graph contrastive learning using the three-view data as an example. We first construct a total of $NV$ relation graphs $G^{(v)}_n$ centered on each sample via $K$-Nearest Neighbor (KNN). In partially aligned multi-view data, only aligned samples have inter-view correspondences. In cross-view graph contrastive learning module, to explore the characteristics from multiply views, the aligned samples in the neighbors of each sample are used to construct the relation graph $G^{(v)}_n$. In detail, the constructed relation graph centered on sample $x^{(v)}_n$ can be denoted by $G^{(v)}_n = \{x^{(v)}_{n^1},...,x^{(v)}_{n^K}\}$, where $x^{(v)}_{n^1},...,x^{(v)}_{n^K}$ are the aligned neighboring samples of $x^{(v)}_n$ and $K$ is the number of neighboring samples in the relation graph. Since different source/modality characteristics of the same instance are semantically consistent, the same instance in different views can share a similar relation graph. We then map the aligned neighboring samples in $G^{(v)}_n$ to other views based on the alignment relationship to obtain the cross-view graph $\Tilde{G}^{(v)}_n$. Therefore, the cross-view relation graph of $x^{(v)}_n$ can be denoted by $\Tilde{G}^{(v)}_n = \{x^{(v)}_{n^1},...,x^{(v)}_{n^K}\}_{v=1}^V$. We can maximize the similarity of the representation of $x^{(v)}_n$ and samples in $\Tilde{G}^{(v)}_n$ while minimizing the similarity of other representations. 

In training, for each representation $z^{(v)}_n$, $VK$ positive pairs can be obtained based on whether a sample is in $\Tilde{G}^{(v)}_n$. Given a mini-batch of size $M$, there are $VKM$ pairs for each view, of which $VK(M-1)$ pairs are negative pairs other than positive pairs. To maximize the similarity of positive pairs while minimizing the similarity of negative pairs, the loss for $z^{(v)}_n$ is defined as
\begin{equation}\small\label{eq:cons}
\begin{aligned}
     & Lc_n^{(v)} = \\
     & -\frac{1}{VK}\sum_{i=1}^V\sum_{k=1}^{K} \log\frac{\exp{(s(z^{(v)}_n, z^{(i)}_{n^k}))}}{\sum_{m=1}^M\left [\exp{(s(z^{(v)}_n, z^{(v)}_m))} + \exp{(s(z^{(v)}_n, z^{(i)}_{m^k}))}\right ]}, \\
\end{aligned}
\end{equation}
where $z^{(i)}_{n^k}$ is the representation of aligned neighboring sample in the cross-view relation graph of $z^{(v)}_n$, $z^{(v)}_m$ denotes the representation in the same mini-batch with $z^{(v)}_n$, and $z^{(i)}_{m^k}$ is in the cross-view relation graph of $z^{(v)}_m$. $s(.)$ is cosine distance to compute the similarity between two representations, which is defined as: 
\begin{equation}
    s(z^{(v)}_i, z^{(v)}_j) = \frac{(z^{(v)}_i)(z^{(v)}_j)^T}{\| z^{(v)}_i \| \| z^{(v)}_j \|},
\end{equation}

To this point, we can implement cross-view graph contrastive learning through Eq.~(\ref{eq:cons}). However, in practice, close samples do not always fall into one category, and the error increases with distance. Thus, there is a certain probability that the neighboring samples in the relation graph $\Tilde{G}^{(v)}_n$ do not belong to the same category as $x^{(v)}_n$. To solve this problem, we sorted the samples in the aligned relation graph by distance, where $x^{(v)}_{n^{1}}$ is the closest sample to $x^{(v)}_n$ in the aligned relation graph $G^{(v)}_n$ and a distance-weighted graph contrastive loss can be derived from Eq.~(\ref{eq:cons}) as follows.


\begin{equation}
    Lw_n^{(v)} = \frac{k^{-1}}{\sum_{k=1}^{K} k^{-1}} Lc_n^{(v)}
\end{equation}
where $k$ is an ordinal index, with neighboring samples of smaller $k$ being closer to the central sample. There are two reasons for using distance index-weighted instead of distance-weighted. Firstly, cosine distance is used to compute the distance matrix in the KNN algorithm. This allows samples to be at different distances from their nearest neighbors in different views, which increases the complexity of computing the loss function and may lead to hard convergence. Secondly, If we use distance-weighted graph comparison loss, we must store the distance of each sample to samples in the relation graphs. This can bring additional space complexity during training.
Then, the cross-view graph contrastive loss is generalized from $z^{(v)}_n$ to the whole dataset, and the final loss can be described as 
\begin{equation}
    L_{CGC} = \frac{1}{NV} \sum_n^N \sum_v^V Lw_n^{(v)} 
\end{equation}

Finally, the overall objective function of CIRCLE can be formulated as:
\begin{equation}\label{eq:all}
    \mathcal{L} = L_{REC} +\lambda L_{CGC}
\end{equation}
where $\lambda$ is a weight parameter.

\begin{algorithm}[t]
\caption{The CIRCLE algorithm}
\begin{algorithmic}
\STATE \textbf{Input:} A partially aligned multi-view dataset $X^{(v)}$, loss weight parameter $\lambda$,  number of aligned neighbors $K$, batch size $M$, maximum iterations $MaxEpoch$.
\STATE \textbf{Output:} The learned common representation $Z$.
\STATE \textbf{Initialization:} Initialize parameters $\theta^{(v)}$ of the view-specific autoencoders.
\STATE $\backslash\backslash$ Training
\STATE Construct relation graphs $G^{(v)}_n$ on each samples.
\STATE Fuse relation graphs $G^{(v)}_n$  to obtain the cross-view relation graph  $\tilde{G}^{(v)}_n$.
\FOR{$epoch = 0$ \TO $MaxEpoch$}
\STATE Sample a mini-batch $\{x^{(v)}_i\}_{i=1}^M$ and their cross-view relation graph  $\tilde{G}^{(v)}_i$ from $X^{(v)}$.
\STATE Input $\{x^{(v)}_i\}_{i=1}^M$ to view-specific autoencoders to obtain the representation $\{z^{(v)}_i\}_{i=1}^M$ and the reconstructed samples $\{\hat{x}^{(v)}_i\}_{i=1}^M$.
\STATE Input the samples in $\tilde{G}^{(v)}_i$ into view-specific autoencoders to obtain $\{z_{i^k}^{(v)}\}_{i=1}^M$.
\STATE Compute the reconstruction loss $L_{REC}$ using Eq.(1). 
\STATE Compute the distance-weighted graph contrastive loss $L_{CGC}$ using Eq.(4) and Eq.(5).
\STATE Compute the overall loss $\mathcal{L}$ by Eq. (6).
\STATE Update $\theta^{(v)}$ to minimize $\mathcal{L}$.
\ENDFOR
\STATE $\backslash\backslash$ Test
\FOR{$x^{(v)}$ \TO $X^{(v)}$}
\STATE Input $x^{(v)}$ into view-specific encoders to obtain $z^{(v)}$.
\ENDFOR
\STATE Calculate the distance matrix between the representations of each view and the correspondences in other views with the smallest distances.
\STATE Concatenate the corresponding representations to obtain $Z$ for downstream tasks.
\RETURN The learned common representation $Z$.
\end{algorithmic}
\end{algorithm}

\subsection{Implementation}

Our model can be performed in three steps to achieve cluster-level alignment and learn latent representation simultaneously.

\textit{\textbf{Step 1.}} The aligned view-specific relation graphs $G^{(v)}$ are constructed via the KNN algorithm contained in the Scikit-Learn package \cite{journals/jmlr/PedregosaVGMTGBPWDVPCBPD11} for all samples in partially aligned multi-view data. We then merge $G^{(v)}$ to obtain cross-view graph $\Tilde{G}^{(v)}$.

\textit{\textbf{Step 2.}} The samples and their aligned neighbor samples are fed into the view-specific autoencoders to learn the latent representations. Then the overall loss can be computed by Eq.~\ref{eq:all}, and the network parameters are updated via back-propagation.

\textit{\textbf{Step 3.}} Once our model converges, we use Euclidean distances to calculate the distance matrix between the representations of each view. For each representation $z_n^{(1)}$ in the first view, its correspondences in other views are representations with the smallest distances. Finally, we simply concatenate the corresponding representations, which are used for downstream tasks such as clustering and classification.

For convenience, the proposed method is summarized in Algorithm 1.

\subsection{Complexity Analysis}

Let $N$, $V$, $K$ denote the data size, the number of views, and the number of neighbors used for graph contrastive learning, respectively. Let $E$ represent the maximum number of neurons in hidden layers of deep neural networks. $H$ denotes the dimensionality of high-level features. $M$ is the size of mini-batch. Generally $V, K, H, M << N, E$ holds. In the mini-batch optimization process, it is not difficult to find that the complexity to compute the reconstruction loss, and cross-view graph contrastive loss is $O(MV)$, and $O(M^3V^2K)$, respectively. The complexity of view-specific autoencoders is $O(MVE^2)$. Therefore, the total complexity to train the model is $O(N/M(M^3V^2K+MVE^2))$, , which is linear to $N$. In conclusion, the complexity of CIRCLE is linear to the data size $N$ which can be easily applied to large-scale data tasks.

\section{Experiments}
\label{SECExperiments}
In this section, we conduct experiments on five widely-used multi-view datasets to evaluate the representation learned by CIRCLE through clustering and classification tasks. 

\subsection{Experimental Settings}

\begin{table}[t]
\centering
\caption{Dataset statistics}
\resizebox{8.5cm}{!}{
\begin{tabular}{lcccccc}
\toprule
Datasets    & Classes & Views  & Instances  & Features \\
\midrule
Scene-15    & 15      & 2      & 4485       & 59/20\\
Caltech-101 & 101     & 2      & 9144       & 1984/512\\
Reuters     & 6       & 2      & 18758      & 10/10\\
BBCSport    & 5       & 3      & 282        & 2582/2544/2465\\
100Leaves   & 100     & 3      & 1600       & 64/64/64\\
\bottomrule
\end{tabular}}\label{tab:data}
\end{table}

\begin{table*}[t]
\centering
\caption{Clustering results on five multi-view datasets, where "$-$" indicates out of memory and partially aligned multi-view representation learning or clustering methods are \underline{underlined}. Since some methods cannot handle data with more than two views, their results are obtained from the best two views of BBCSport and 100Leaves.}\label{tab:cluster_compare}
\resizebox{\textwidth}{!}{
\begin{tabular}{l|l|l|ccc|ccc|ccc|ccc|ccc}
\toprule
\multirow{2}{*}{Aligned}    & \multirow{2}{*}{Methods} & Year/       & \multicolumn{3}{c|}{Scene-15} & \multicolumn{3}{c|}{Caltech-101} & \multicolumn{3}{c|}{Reuters} & \multicolumn{3}{c|}{BBCSport} & \multicolumn{3}{c}{100Leaves} \\
                            &                          & Publication & ACC      & NMI     & ARI     & ACC       & NMI      & ARI      & ACC     & NMI     & ARI     & ACC      & NMI     & ARI     & ACC      & NMI      & ARI     \\
\midrule
\multirow{9}{*}{Fully}      & DCCA                     & 2013/ICML   & 36.61    & 39.20   & 21.03   & 27.60     & 47.84    & 30.86    & \textbf{47.95}   & 26.57   & 12.71   & 70.57    & 62.09   & 52.59   & 61.44    & 81.62    & 49.26   \\
                            & DCCAE                    & 2015/ICML   & 34.58    & 39.01   & 19.65   & 19.65     & 45.05    & 14.57    & 41.98   & 20.30   & 8.51    & 71.28    & 59.47   & 54.99   & 61.25    & 80.94    & 48.06   \\
                            & MVGL                     & 2018/TCYB   & 32.04    & 36.64   & 21.54   & 25.37     & 33.25    & 0.57     & $-$     & $-$     & $-$     & 37.59    & 6.00    & 0.80    & 81.06    & 89.12    & 51.55   \\
                            & MCGC                     & 2019/TIP    & 21.05    & 21.33   & 11.81   & 32.03     & 37.71    & 3.80     & 27.01   & 2.20    & 0.10    & 39.00    & 8.64    & 0.94    & 76.19    & 85.79    & 51.86   \\
                            & GMC                      & 2019/TKDE   & 24.97    & 29.56   & 4.03    & 20.21     & 32.61    & 0.30     & 30.88   & 8.95    & 0.57    & \textbf{88.65}    & \textbf{79.97}   & \textbf{78.99}   & \textbf{83.19}    & \textbf{93.09}    & 52.45   \\
                            & AE$^2$-Nets              & 2019/CVPR   & 37.17    & \textbf{40.47}   & 22.24   & 20.79     & 45.01    & 15.89    & 42.39   & 19.76   & 14.87   & 41.84    & 18.33   & 4.87    & 75.00    & 88.80    & \textbf{67.14}   \\
                            & LMVSC                    & 2020/AAAI   & 35.12	& 35.58	  & 18.92   & 27.62	    & 47.39    & 20.77    & 45.93   & 17.68   & 17.19   & 74.11    & 54.21   & 51.46   & 70.38    & 86.31    & 60.53   \\
                            & OPMC                     & 2021/ICCV   & \textbf{37.26}    & 40.40   & 21.55   & 28.46     & \textbf{50.38}    & 26.76    & 45.68	& \textbf{21.75}	  & 15.10   & 72.34    & 59.68   & 57.56   & 65.12    & 85.22    & 59.35   \\
                            & SMVSC                    & 2021/MM     & 36.76	& 35.33   &	\textbf{25.98}   & \textbf{36.26}     & 47.69	   & \textbf{37.85}    & 42.52	& 18.91	  & \textbf{33.18}   & 43.26    & 14.22	 & 35.37   & 37.50    & 71.17	 & 25.65   \\
\midrule
\multirow{13}{*}{Partially} & DCCA                     & 2013/ICML   & 34.27    & 36.55   & 18.83   & 12.52     & 32.13    & 7.63     & 39.71   & 13.83   & 14.38   & 44.33    & 6.72    & 6.99    & 46.81    & 72.12    & 31.38   \\
                            & DCCAE                    & 2015/ICML   & 33.62    & 36.56   & 18.54   & 11.75     & 30.54    & 6.60     & 41.42   & 12.82   & 13.61   & 36.17    & 6.09    & 1.82    & 47.44    & 73.08    & 32.29   \\
                            & MVGL                     & 2018/TCYB   & 30.24    & 34.37   & 15.58   & 14.66     & 16.74    & 0.71     & $-$     & $-$     & $-$     & 36.88    & 5.05    & 1.18    & 62.00    & 77.88    & 27.90   \\
                            & MCGC                     & 2019/TIP    & 13.89    & 4.33    & 2.75    & 24.91     & 25.33    & 1.79     & 26.18   & 2.00    & 0.01    & 38.65    & 4.86    & 0.72    & 39.13    & 50.00    & 3.26    \\
                            & GMC                      & 2019/TKDE   & 11.08    & 4.51    & 0.10    & 15.08     & 21.10    & 0.23     & 26.98   & 4.50    & 0.46    & 54.26    & 25.82   & 20.79   & 38.81    & 56.38    & 2.04    \\
                            & AE$^2$-Nets              & 2019/CVPR   & 28.56    & 26.58   & 12.96   & 10.45     & 29.51    & 7.90     & 35.49   & 10.61   & 8.07    & 36.17    & 4.84    & 0.61    & 49.81    & 73.85    & 34.43   \\
                            & LMVSC                    & 2020/AAAI   & 27.76	& 19.03	  & 10.89   & 20.99	    & 38.95    & 15.11    & 37.02   & 9.16    & 10.17   & 53.19    & 22.82   & 15.67   & 43.94    & 62.26    & 16.20   \\
                            & OPMC                     & 2021/ICCV   & 29.90    & 27.15   & 12.55   & 25.34     & 47.57    & 21.87    & 37.48   & 10.62   & 7.77    & 35.11    & 7.63    & 4.84    & 21.12    & 51.32    & 8.63    \\
                            & SMVSC                    & 2021/MM     & 22.92	& 13.96	  & 15.47   & 29.43	    & 41.41    & 31.92    & 38.13	& 11.40   & 25.42   & 36.52    & 62.07   & 31.86   & 21.44    &	49.66	 & 9.17    \\
                            & \underline{PVC}          & 2020/NeurIPS& 37.88    & 39.12   & 20.63   & 22.11     & 47.82    & 17.98    & 42.07   & 20.43   & 16.95   & 45.03    & 19.25   & 14.07   & 39.56    & 63.76    & 22.22   \\                
                            & \underline{MVC-UM}       & 2021/KDD    & 25.70 	& 27.70	  & 11.54   & 11.20	    & 33.57	   & 9.08     & 34.02   & 11.10   & 12.25   & 44.31    & 17.69   & 12.91   & 54.88	  & 79.43	 & 41.75   \\
                            & \underline{MvCLN}        & 2021/CVPR   & 38.53    & 39.90   & 24.26   & 30.09     & 43.07    & \textbf{38.34}    & 50.16   & 30.65   & 24.90   & 40.43    & 12.41   & 8.25    & 40.19    & 70.39    & 26.49   \\
                            & \underline{CIRCLE}       & Ours        & \textbf{45.24}    & \textbf{44.39}	  & \textbf{27.13}   & \textbf{30.30}     & \textbf{50.64}	   & 32.10    & \textbf{51.81}   & \textbf{31.98}   & \textbf{26.61}   & \textbf{85.11}    & \textbf{68.28}   & \textbf{70.16}   & \textbf{81.25}    & \textbf{88.51}	 & \textbf{69.71}   \\
\bottomrule
\end{tabular}}
\end{table*}

\subsubsection{Datasets}
Five widely-used datasets (listed in Table. ~\ref{tab:data}) are used to evaluate the proposed method. 
(1) \textbf{Scene-15} \cite{conf/cvpr/LiPT05} contains 4485 images of 15 categories with two image features as views.
(2) \textbf{Caltech-101} \cite{journals/pami/ZhangLSSS19} is a widely-used image dataset containing 9144 instances from 101 categories. Following \cite{conf/cvpr/YangL0L0021}, the 1,984-dim HOG feature and 512-dim GIST feature are used as two views.
(3) \textbf{Reuters} \cite{conf/nips/AminiUG09} consists of 18758 instances of six subjects, and English and French are used as two views. 
(4) \textbf{BBCSport} \cite{conf/kdd/YuTWG21} is a synthetic multi-view text dataset from BBC Sport corpora. We follow \cite{conf/kdd/YuTWG21} to use 282 samples from 5 classes. 
(5) \textbf{100Leaves} \cite{conf/ijcai/WangZLYZ19} contains 1600 instances from 100 categories, and shape descriptor, fine scale margin, and texture histogram features are used as three views.

\subsubsection{Compared methods}
In this section, we compare CIRCLE with 12 multi-view representation learning or clustering baselines, including Deep Canonical Correlation Analysis (DCCA) \cite{conf/icml/AndrewABL13}, Deep Canonical Correlation Analysis Autoencoder (DCCAE) \cite{conf/icml/WangALB15}, Multiview Graph Learning (MVGL) \cite{journals/tcyb/ZhanZGW18}, Multiview Consensus Graph Clustering (MCGC) \cite{journals/tip/ZhanNWY19}, Graph-based Multi-view Clustering (GMC) \cite{journals/tkde/WangYL20}, Autoencoder in Autoencoder Networks (AE$^2$-NET) \cite{conf/cvpr/ZhangLF19}, Large-scale Multi-view Subspace Clustering (LMVSC) \cite{conf/aaai/KangZZSHX20}, One-pass Multi-view Clustering (OPMC) \cite{/conf/iccv/opmc}, Scalable Multi-view Subspace Clustering (SMVSC) \cite{conf/mm/SunZWZTLZW21}, Partially View-aligned Clustering (PVC) \cite{conf/nips/Huang0ZL020}, Multi-view Clustering Method for Unknown Mapping Relationships (MVC-UM) \cite{conf/kdd/YuTWG21}, and Multi-view Contrastive Learning with Noise-robust Loss (MvCLN) \cite{conf/cvpr/YangL0L0021}.

\subsubsection{Evaluation methods and metrics}
To evaluate the clustering performance of the learned representations, we use the K-means++ \cite{conf/soda/ArthurV07} contained in the Scikit-Learn package \cite{journals/jmlr/PedregosaVGMTGBPWDVPCBPD11} with the default configurations. Three common-used statistical metrics, i.e., Accuracy (ACC), Normalized Mutual Information (NMI), and Adjusted Rand Index (ARI), are used to assess the clustering performance. Then, the SVM classifier \cite{Platt99probabilisticoutputs} contained in the Scikit-Learn package is used to evaluate the classification performance. We divide the learned representations into a training set and a test set with different proportions, and the training set is used to train the SVM classifier. After training, we use the test set to verify performance and report classification accuracy. For all the metrics, a higher value indicates a better performance.

\subsubsection{Partially aligned multi-view data construction}
To evaluate the performance on partially aligned multi-view data, the first view is regarded as the baseline view and randomly disrupts 50\% of the samples in other views. There are only four methods, i.e., PVC, MVC-UM, MvCLN, and our CIRCLE can handle partially aligned multi-view data. To be fair, for the other nine methods, PCA is first used to map the original multi-source data to the same dimension, then the Hungarian algorithm is used to build the correspondence, and these methods are performed on the re-aligned multi-view data finally. As a comparison, we also report the results of these baselines on the original alignment dataset. To avoid the effects of randomness on performance, we run all methods five times and report the average metrics.

\subsection{Experimental Results and Analyses}

\begin{figure}[t]
    \centering
    \includegraphics[width=8cm]{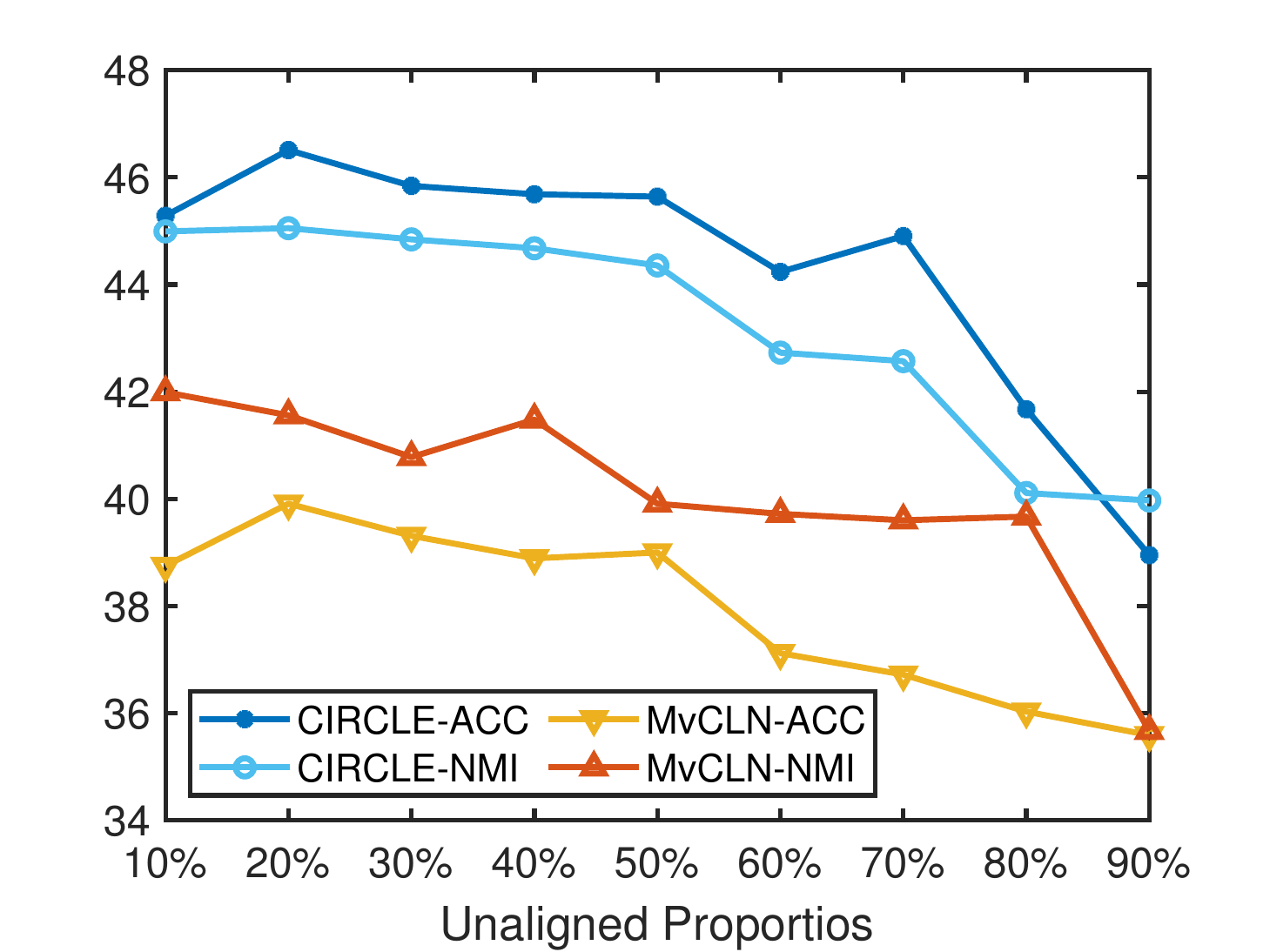}
    \caption{Clustering performance comparison with different unaligned proportions}
    \label{fig:differalign}
\end{figure}

\begin{table*}[t]
\centering
\caption{Classification accuracy results on five multi-view datasets, where "$-$" indicates out of memory and partially aligned multi-view representation learning or clustering methods are \underline{underlined}. The proportions of the training set used to train the SVM classifier are set to 80\%, 50\%, and 20\%, and the rest of the data is used to validate the classification performance. Note: since OPMC \cite{/conf/iccv/opmc} directly obtains the clustering assignments, we cannot obtain the representation used for the classification task.}\label{tab:class_compare}
\resizebox{\textwidth}{!}{
\begin{tabular}{l|l|l|ccc|ccc|ccc|ccc|ccc}
\toprule
\multirow{2}{*}{Aligned}    & \multirow{2}{*}{Methods} & Year/       & \multicolumn{3}{c|}{Scene-15} & \multicolumn{3}{c|}{Caltech-101} & \multicolumn{3}{c|}{Reuters} & \multicolumn{3}{c|}{BBCSport} & \multicolumn{3}{c}{100Leaves} \\
                            &                          & Publication & 80\%\tnote{1}& 50\%    & 20\%    & 80\%      & 50\%    & 20\%      & 80\%    & 50\%    & 20\%    & 80\%     & 50\%    & 20\%    & 80\%     & 50\%    & 20\%    \\
\midrule
\multirow{8}{*}{Fully}      & DCCA                     & 2013/ICML   & 63.61    & 61.72   & 57.30   & 38.89     & 37.23    & 33.75    & 71.92   & 72.33   & 71.54   & 85.71    & 80.78   & 80.44   & 80.62    & 76.12    & 63.75   \\
                            & DCCAE                    & 2015/ICML   & 50.42    & 48.84   & 46.48   & 38.61     & 37.53    & 34.03    & 72.00   & 71.65   & 70.63   & 87.50    & 86.65   & 80.22   & 78.13    & 71.38    & 59.14   \\
                            & MVGL                     & 2018/TCYB   & 52.95    & 52.81   & 50.25   & 56.46     & 54.02    & 47.31    & $-$     & $-$     & $-$     & 58.93    & 45.39   & 34.22   & 92.50    & 87.75    & 86.48   \\    
                            & MCGC                     & 2019/TIP    & 42.47    & 41.88   & 39.72   & 46.94     & 46.87    & 42.88    & 58.10   & 57.99   & 52.60   & 48.21    & 40.43   & 38.22   & 87.19    & 84.75    & 80.00   \\
                            & GMC                      & 2019/TKDE   & 62.88    & 60.48   & 56.55   & \textbf{60.94}     & \textbf{58.60}    & \textbf{53.85}    & \textbf{83.66}   & \textbf{81.91}   & \textbf{74.56}   & \textbf{91.07}    & \textbf{90.91}   & \textbf{80.89}   & \textbf{96.25}    & \textbf{95.50}    & \textbf{88.12}   \\
                            & AE$^2$-Nets              & 2019/CVPR   & \textbf{72.03}    & \textbf{69.76}   & \textbf{64.66}   & 35.24     & 34.38    & 31.72    & 65.47   & 64.82   & 63.28   & 85.71    & 80.85   & 71.11   & 90.62    & 82.37    & 65.47   \\
                            & LMVSC                    & 2020/AAAI   & 49.28    & 48.84   & 43.20   & 51.70     & 50.74    & 47.74    & 52.31   & 52.37   & 50.85   & 89.29    & 76.60   & 77.78   & 93.56    & 90.75    & 69.69   \\
                            & SMVSC                    & 2021/MM     & 38.57    & 37.20   & 37.43   & 42.45     & 42.61    & 40.07    & 52.57   & 52.39   & 49.08   & 41.07    & 41.13   & 38.67   & 32.19    & 31.87    & 27.58   \\

\midrule
\multirow{12}{*}{Partially} & DCCA                     & 2013/ICML   & 51.68    & 50.64   & 46.85   & 35.72     & 33.97    & 31.20    & 65.92   & 65.80   & 65.15   & 46.43    & 41.06   & 37.78   & 53.12    & 52.75    & 35.86   \\
                            & DCCAE                    & 2015/ICML   & 46.24    & 45.37   & 43.75   & 31.95     & 30.75    & 28.14    & 61.88   & 61.58   & 60.67   & 46.43    & 42.55   & 38.67   & 51.88    & 46.50    & 32.81   \\
                            & MVGL                     & 2018/TCYB   & 54.52    & 50.40   & 50.22   & 45.79     & 44.12    & 37.79    & $-$     & $-$     & $-$     & 39.29    & 38.30   & 35.11   & 69.37    & 67.87    & 62.73   \\
                            & MCGC                     & 2019/TIP    & 35.65    & 32.57   & 30.33   & 41.41     & 40.16    & 36.21    & 48.44   & 47.74   & 37.20   & 48.49    & 47.37   & 40.89   & 61.56    & 57.25    & 44.06   \\
                            & GMC                      & 2019/TKDE   & 50.72    & 47.64   & 41.75   & 50.93     & 48.27    & 41.95    & 71.98   & 67.37   & 56.56   & 62.50    & 61.70   & 46.22   & 57.81    & 57.13    & 42.89   \\
                            & AE$^2$-Nets              & 2019/CVPR   & 48.19    & 47.64   & 42.61   & 23.30     & 22.65    & 20.61    & 62.74   & 62.40   & 60.65   & 42.86    & 41.13   & 37.78   & 74.69    & 62.62    & 47.73   \\
                            & LMVSC                    & 2020/AAAI   & 35.12    & 36.98   & 30.07   & 49.51     & 47.31    & 44.65    & 48.63   & 48.28   & 46.35   & 53.57    & 55.32   & 49.78   & 66.25    & 59.50    & 47.73   \\
                            & SMVSC                    & 2021/MM     & 26.20    & 25.20   & 23.47   & 37.64     & 38.34    & 35.87    & 44.44   & 42.96   & 42.67   & 39.29    & 34.04   & 21.33   & 13.75    & 15.88    & 10.16   \\
                            & \underline{PVC}          & 2020/NeurIPS& 48.77    & 45.97   & 40.46   & 36.78     & 36.50    & 35.54    & 72.63   & 72.08   & 71.11   & 66.07    & 53.90   & 54.22   & 46.56    & 39.37    & 24.14   \\
                            & \underline{MVC-UM}       & 2021/KDD    & 32.60    & 32.67   & 31.15   & 31.48     & 30.38    & 27.19    & 63.04   & 62.19   & 60.01   & 47.50    & 42.44   & 28.37   & 53.12    & 52.62    & 43.12   \\
                            & \underline{MvCLN}        & 2021/CVPR   & 57.93    & 57.15   & 55.52   & 46.69     & 45.89    & 43.87    & \textbf{81.77}   & \textbf{81.63}   & \textbf{81.11}   & 50.00    & 40.43   & 31.11   & 58.67    & 57.05    & 51.43   \\
                            & \underline{CIRCLE}                   & Ours        & \textbf{71.01}    & \textbf{69.68}   & \textbf{65.86}   & \textbf{62.25}     & \textbf{58.77}    & \textbf{49.84}    & 81.42   & 81.17   & 80.44   & \textbf{91.07}    & \textbf{90.78}   & \textbf{83.56}   & \textbf{85.31}    & \textbf{81.25}    & \textbf{74.22}   \\

\bottomrule
\end{tabular}}
\end{table*}

\subsubsection{Clustering comparisons}
Table. ~\ref{tab:cluster_compare} shows the clustering results of CIRCLE and baseline methods. The mean metrics are given in the table, with the best results are highlighted in \textbf{bold}. Since some methods cannot be applied to data with more than two views, we report their results on the best two views of BBCSport and 100Leaves datasets. From the experimental results, we have the following observations: (1) The proposed CIRCLE achieves the best performance on all partially aligned datasets with most evaluation metrics, validating the effectiveness of the proposed method. (2) CIRCLE can still achieve competitive performance compared to baselines performed on aligned datasets, especially on the Scene-15 dataset, where CIRCLE can achieve about 8\% ACC improvement compared to the best baseline. (3) Although the proposed CIRCLE improves modestly compared to MvCLN on the two-view dataset, it has a significant improvement on three-view datasets like BBCSport. Since MvCLN can only perform on data with two views, even with the best two views of the three-view dataset, it is still difficult to explore comprehensive information.

\begin{figure*}[t]
\centering
\subfloat[PVC]
{
\includegraphics[width=4.3cm]{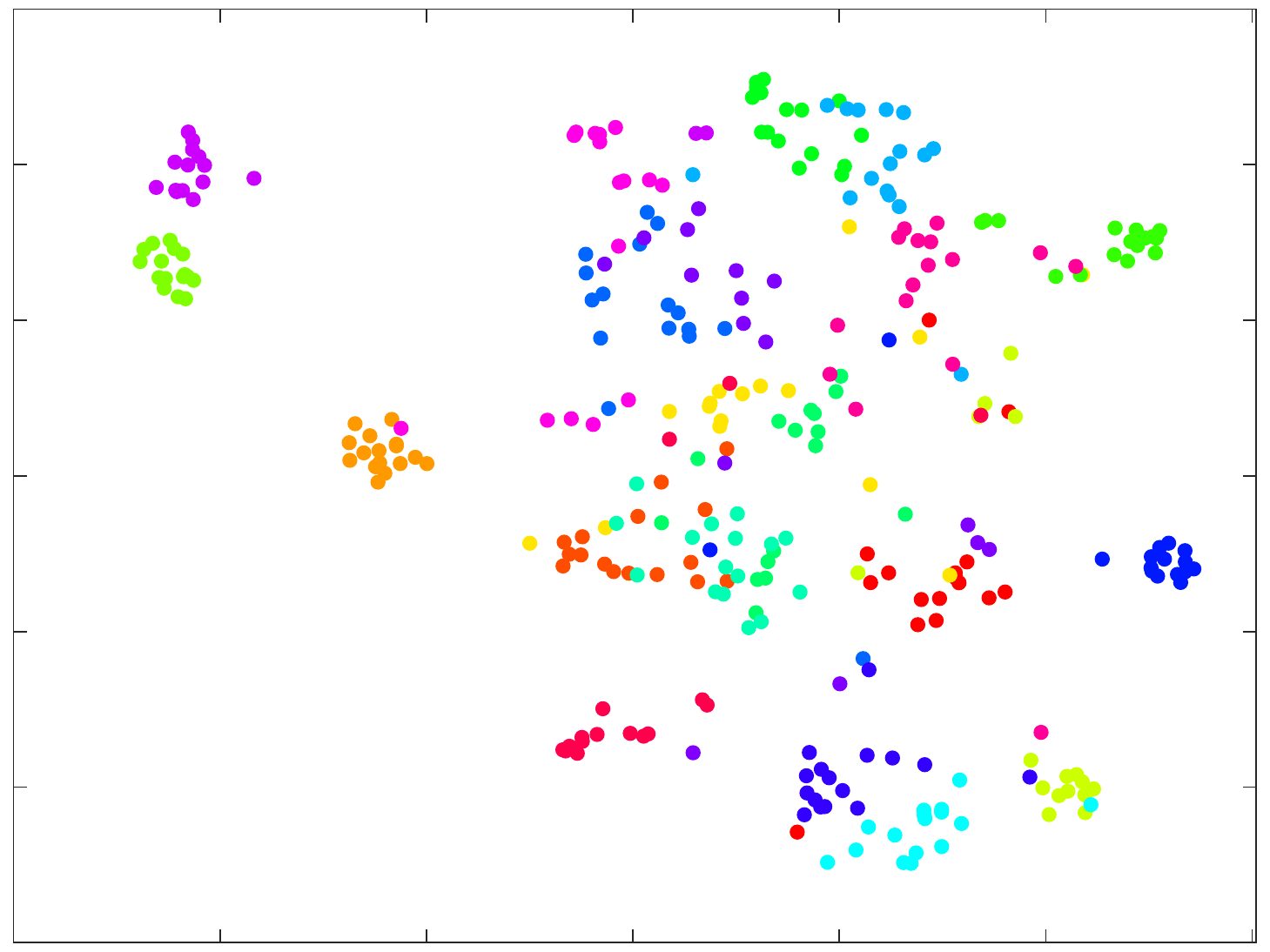}
}\hspace{-2mm}
\subfloat[MVC-UM]
{
\includegraphics[width=4.3cm]{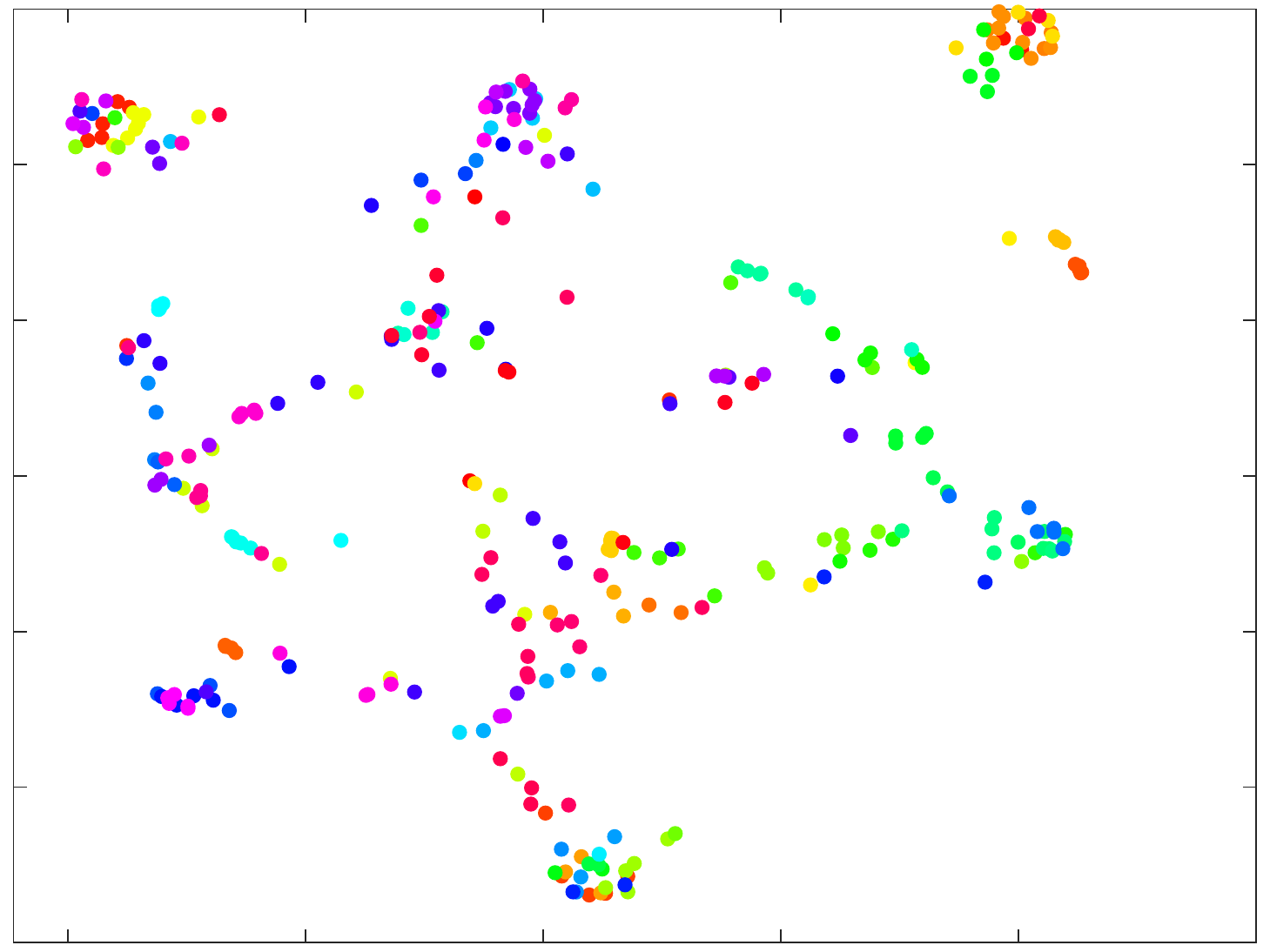}
}\hspace{-2mm}
\subfloat[MvCLN]
{
\includegraphics[width=4.3cm]{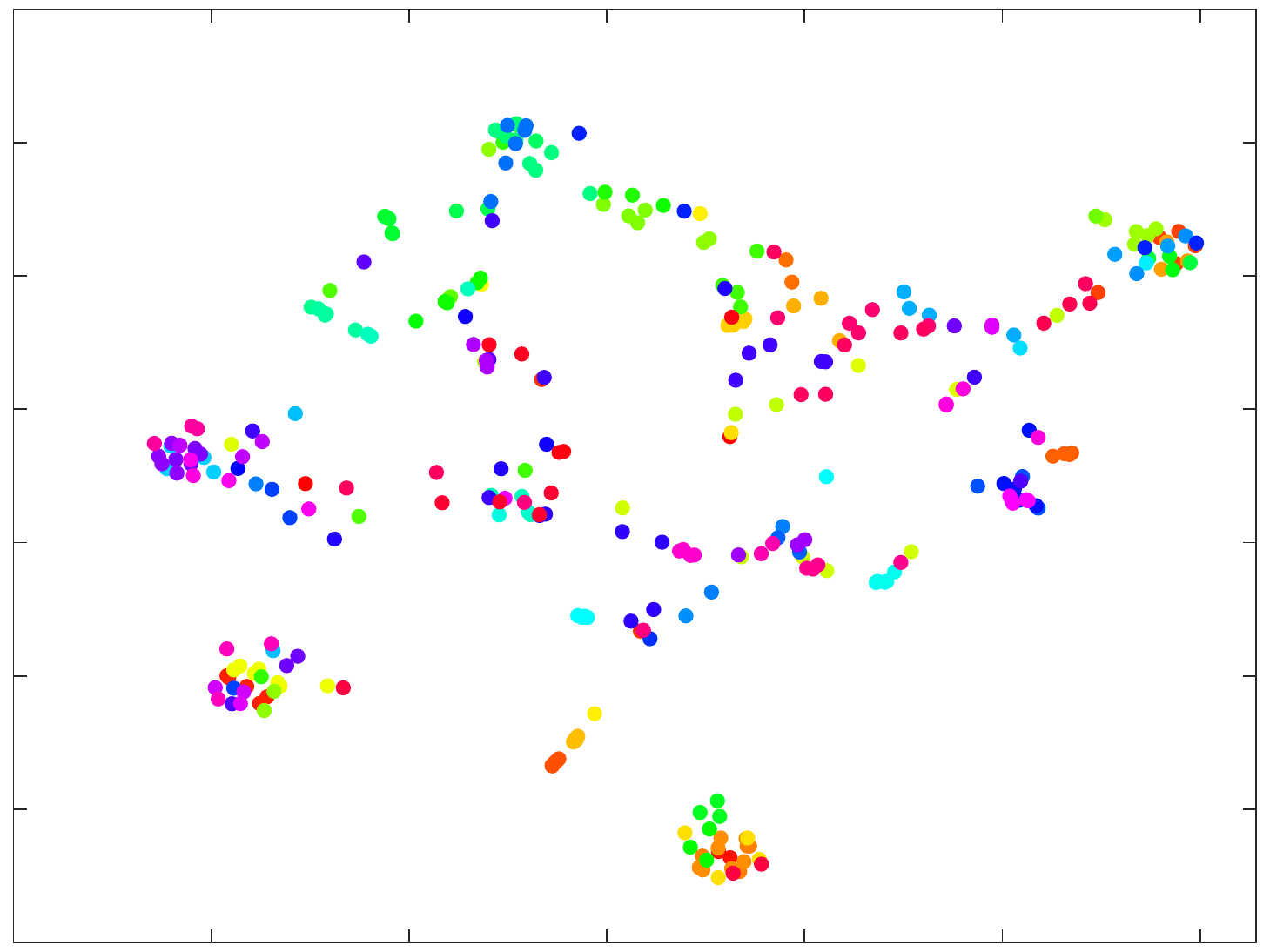}
}\hspace{-2mm}
\subfloat[CIRCLE]
{
\includegraphics[width=4.3cm]{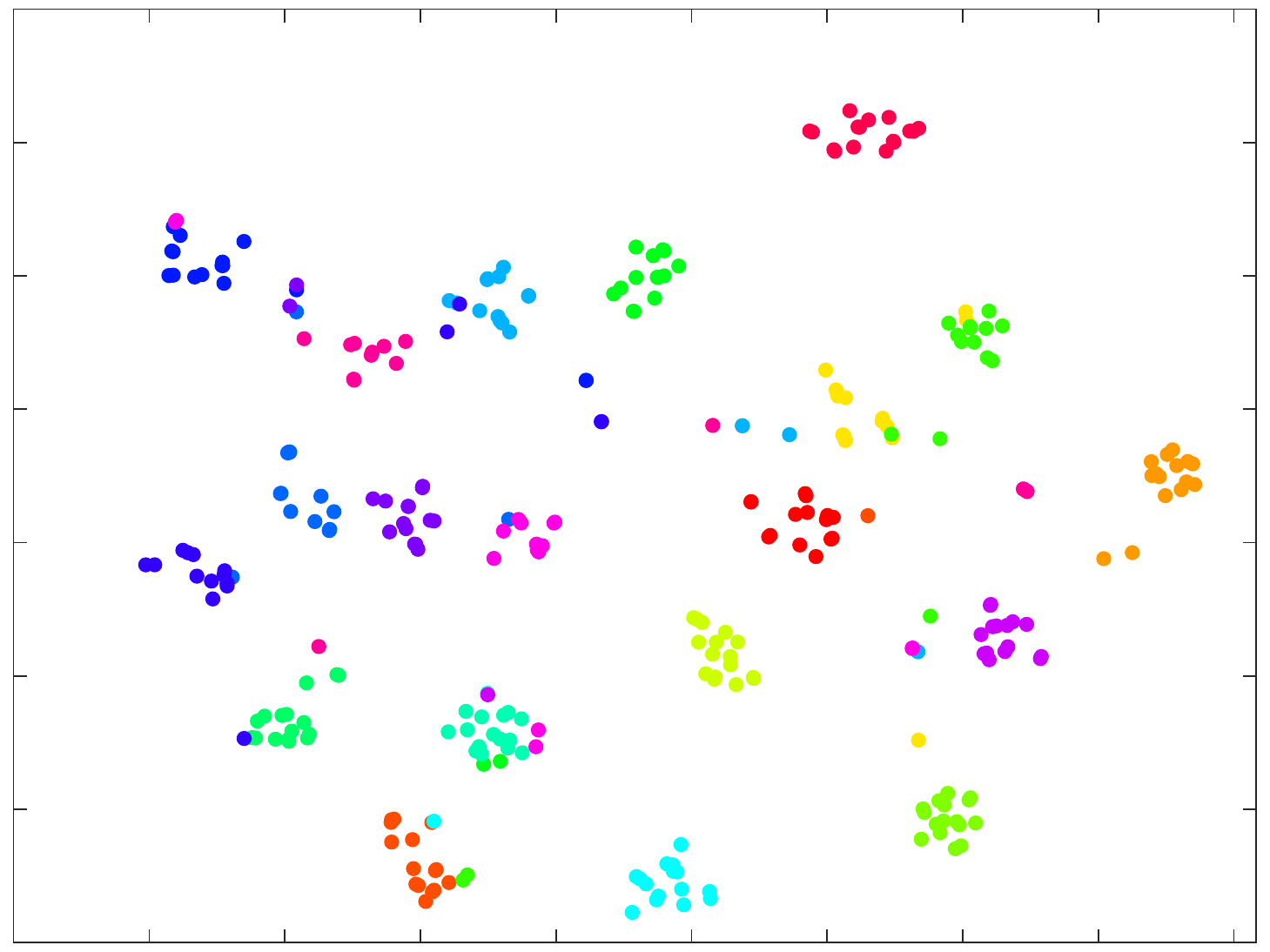}
}
\caption{$t$-SNE visualizations of representations learned by different partially aligned multi-view learning approaches.}
\label{fig:vis}
\end{figure*}

\subsubsection{Clustering comparisons with different aligned proportions}
From Table. ~\ref{tab:cluster_compare}, we notice that deep partially aligned multi-view representation learning methods MvCLN and CIRCLE have similar high performance. For a more careful comparison, we increase the unaligned proportion from 10\% to 90\% with an interval of 10\%, and the clustering performance on the Scene-15 dataset is reported in Fig. ~\ref{fig:differalign}. From the results, we can observe that: (1) Compared with the best partially aligned baseline MvCLN, the proposed CIRCLE shows significant improvement in various unaligned proportions, which demonstrates the superiority of the proposed method. (2) The proposed CIRCLE can obtain stable results until the proportion of unaligned instances reaches 50\%, while MvCLN starts to deteriorate when the proportion of unaligned instances reaches 40\%, which indicates that CIRCLE can learn aligned mappings and discriminative representations with fewer aligned instances.

\subsubsection{Classification comparisons}
We further validate the performance of the learned representations on the classification task through the SVM classifier \cite{Platt99probabilisticoutputs}, where the average accuracy results are shown in Table.~\ref{tab:class_compare}. Since OPMC obtains the clustering assignment directly, it cannot be used for the classification task. For graph-based methods, i.e., MVGL, MCGC, and GMC, spectral representations are used for classification. For other methods, the learned common representations or concatenated view-specific representations are used for classification. We can observe the following points from Table.~\ref{tab:class_compare}: (1) The proposed CIRCLE achieves the best classification performance on the most partially aligned dataset with various training/testing set proportions, validating the effectiveness of the learned representations on the classification task. (2) Compared to fully aligned baselines, the proposed CIRCLE still achieves competitive results though the baselines are with ground-truth alignment. CIRCLE is even better than all the baselines on the Caltech-101 and BBCSport datasets, which demonstrates that the representations learned by the cluster-level alignment are valid for the classification task.

\subsubsection{Visualization}
In order to show the superiority of the representation obtained by CIRCLE, we visualize representations learned by different partially aligned multi-view learning approaches on 100Leaves with $t$-SNE \cite{jmlr/laurens08}. The results are shown in Fig.~\ref{fig:vis}, and the representation of the first twenty categories is visualized for ease of viewing. As can be seen from Fig.~\ref{fig:vis}, MVC-UM and MvCLN cannot recover the structure of the data, mixing each category of data together. PVC can detach the representation of some categories in latent space. In contrast, the proposed CIRCLE can recover a better structure of data since it has smaller intra-cluster scatter and larger inter-cluster scatter. 

\begin{table*}[t]
\centering
\caption{Time cost comparisons. The best result for each setting is in bold and "$-$" indicates the method does not involve this phase. Note: PVC and MvCLN can only handle two views simultaneously, and the table reports the time consumption of PVC and MvCLN performing once on the two selected views.}\label{tab:timecost}
\resizebox{\textwidth}{!}{
\begin{tabular}{c|ccc|ccc|ccc|ccc|ccc}
\toprule
 \multirow{3}{*}{Method}    & \multicolumn{3}{c|}{Scene-15}               & \multicolumn{3}{c|}{Caltech-101}            & \multicolumn{3}{c|}{Reuters}                & \multicolumn{3}{c|}{BBCSport}               & \multicolumn{3}{c}{100Leaves}               \\
                            & Training & Inferring & \multirow{2}{*}{ACC $\uparrow$} & Training & Inferring & \multirow{2}{*}{ACC $\uparrow$} & Training & Inferring & \multirow{2}{*}{ACC $\uparrow$} & Training & Inferring & \multirow{2}{*}{ACC $\uparrow$} & Training & Inferring & \multirow{2}{*}{ACC $\uparrow$} \\
                            & Time (s) $\downarrow$ & Time (s) $\downarrow$  &                      & Time (s) $\downarrow$ & Time (s) $\downarrow$ &                      & Time (s) $\downarrow$ & Time (s) $\downarrow$ &                      & Time (s) $\downarrow$ & Time (s) $\downarrow$  &                      & Time (s) $\downarrow$ & Time (s) $\downarrow$ &                      \\
\midrule
PVC                         & 8834.68  & 1.65      & 37.88                & 9826.37  & 6.05      & 22.11                & 15908.04 & 26.11     & 42.07                & 688.41   & 0.85      & 45.03                & 2800.19  & 1.21      & 39.56                 \\
MVC-UM                      & $-$      & 2897.45   & 25.70                & $-$      & 23647.15  & 11.20                & $-$      & 214186.39 & 34.02                & $-$      & 23.32     & 44.31                & $-$      & 499.47    & 54.88                 \\
MvCLN                       & \textbf{132.12}   & 0.70      & 38.53                & \textbf{315.36}   & 1.38      & 30.09                & \textbf{620.03}   & 3.17      & 50.16                & \textbf{20.31}    & \textbf{0.21}      & 40.43                & \textbf{67.81}    & \textbf{0.57}      & 40.19                 \\
CIRCLE                      & 540.08   & \textbf{0.68}      & \textbf{45.24}                & 1222.05  & \textbf{0.75}      & \textbf{30.30}                & 3031.20  & \textbf{0.86}      & \textbf{51.81}                & 105      & 0.54      & \textbf{85.11}                & 485.97   & 0.68      & \textbf{81.25}                 \\
\bottomrule
\end{tabular}}
\label{tab:time}
\end{table*}

\begin{figure}[t]
    \centering
    \includegraphics[width=8cm]{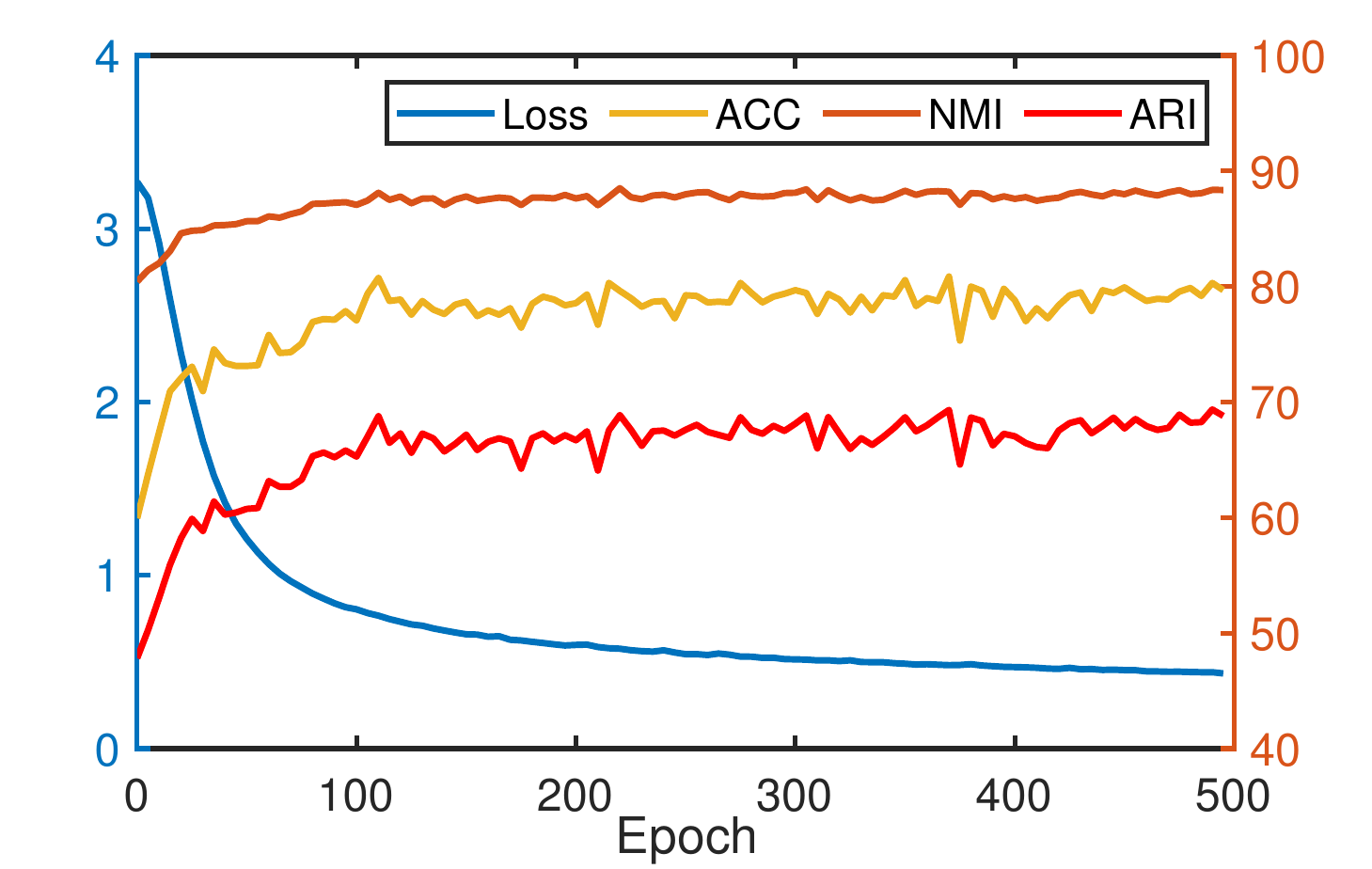}
    \caption{Convergence curve of CIRCLE on 100Leaves dataset. The left and right y-axis denote the loss value and clustering results, respectively.}
    \label{fig:training}
\end{figure}

\subsection{Time cost comparison}

In this section, we quantitatively compare our CIRCLE with other partially aligned multi-view learning approaches PVC, MVC-UM, and MvCLN on the time cost. Table.~\ref{tab:time} shows the training time, inferring time, and clustering ACC of these methods on all datasets. Since the conventional method MVC-UM directly infers the clustering assignments, we use "$-$" to fill the training time. From the results, one could observe the following points: (1) The proposed CIRCLE obtains the shortest inferring time on all two-view datasets and is slightly longer than MvCLN on three-view datasets. Since PVC and MvCLN can only handle two views, the best two are chosen as input when handling a multiply view dataset. However, for a dataset with $V$ views, we need to feed two views into the model in turn for a total of $C_V^2$ runs to select the best two views. This results in a significant increase in the total training and inference time. The table reports the time consumption of PVC and MvCLN performing once on the two selected views, where the actual running time is the value in the table multiplied by $C_3^2$. (2) Due to the introduction of cross-view graph contrastive learning, CIRCLE has a longer training time than MvCLN but is still significantly less than the other two methods. Undeniably, the introduction of cross-view graph contrastive learning also brings a significant performance improvement.

\subsection{Ablation Studies and Convergence Analyses}

\begin{figure}[t]
    \centering
    \includegraphics[width=7.5cm]{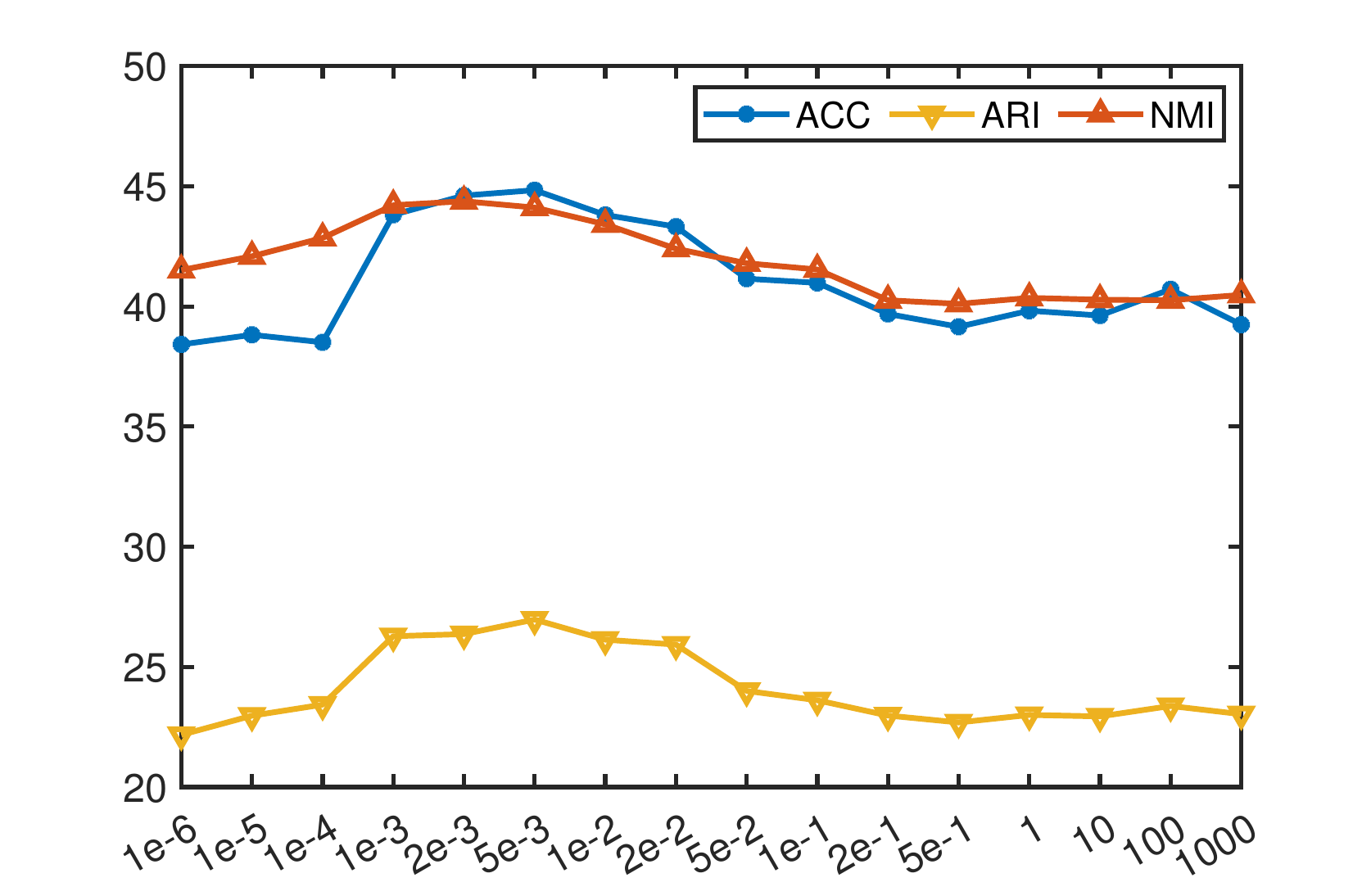}
    \caption{Parameter analysis of $\lambda$ on Scene-15 dataset}
    \label{fig:lambda}
\end{figure}

\textit{\textbf{Ablation studies.}} We further conduct a series of experiments on all used datasets to investigate the effectiveness of diverse components in our model. Table.~\ref{tab:ablation} shows the clustering results on all used datasets for three approaches:  CIRCLE using only $L_{REC}$, CIRCLE using only $L_{CGC}$, and the whole CIRCLE. From the results, we have the following observations: (1) The whole CIRCLE outperforms the other variants, which demonstrates the validity of view-specific reconstruction and cross-view graph contrastive learning. (2) Empirically, reconstruction loss is essential as the basis for autoencoders. However, on most datasets, CIRCLE using only $L_{CGC}$ outperforms $L_{REC}$. In particular, CIRCLE using only $L_{CGC}$ gains more than 30\% improvements in all metrics on the BBCSport dataset, which indicates that cross-view graph contrastive learning plays a crucial role in learning representations on partially aligned multi-view data.

\textit{\textbf{Convergence analysis.}} We investigate the convergence of the proposed CIRCLE by reporting the loss value and the corresponding clustering performance with increasing epochs. As shown in Fig.~\ref{fig:training}, we can find that all three metrics increase obviously in the first 100 epochs and then reaches a stable point. Meanwhile, the training loss calculated via Eq.~\ref{eq:all} remarkably decreases in the first 100 epochs and then tends to stabilize after 400 epochs. 

\begin{table*}[t]
\centering
\caption{Ablation study on all used datasets. Different methods use modules identified by "\Checkmark".}\label{tab:ablation}
\resizebox{\textwidth}{!}{
\begin{tabular}{cc|ccc|ccc|ccc|ccc|ccc}
\toprule
\multirow{2}{*}{$L_{REC}$} & \multirow{2}{*}{$L_{CGC}$} & \multicolumn{3}{c|}{Scene-15} & \multicolumn{3}{c|}{Caltech-101} & \multicolumn{3}{c|}{Reuters} & \multicolumn{3}{c|}{BBCSport} & \multicolumn{3}{c}{100Leaves} \\
                           &                            & ACC      & NMI     & ARI      & ACC       & NMI       & ARI      & ACC     & NMI     & ARI      & ACC      & NMI     & ARI      & ACC      & NMI      & ARI     \\
\midrule
\Checkmark                 &                            & 36.48	   & 40.60	 & 21.09    & 28.25     & 49.59     & 27.42    & 39.54   & 12.77   & 12.81    & 43.97	 & 23.17   & 15.53    & 71.69    & 85.12	& 59.29   \\
                           & \Checkmark                 & 39.38    & 40.75	 & 23.19    & 27.37     & 48.81     & 26.35    & 49.36   & 29.75   & 24.56    & 75.18    & 63.44   & 62.58    & 72.56    & 84.82	& 60.34   \\
\Checkmark                 & \Checkmark                 & \textbf{45.24}    & \textbf{44.39}	 & \textbf{27.13}    & \textbf{30.30}     & \textbf{50.64}	    & \textbf{32.10}    & \textbf{51.81}   & \textbf{31.98}   & \textbf{26.61}    & \textbf{85.11}    & \textbf{68.28}   & \textbf{70.16}    & \textbf{81.25}& \textbf{88.51}	& \textbf{69.71}   \\

\bottomrule
\end{tabular}}
\end{table*}

\begin{figure*}[t]
\centering
\subfloat[$K$]
{
\label{subfig:k}
\includegraphics[width=6cm]{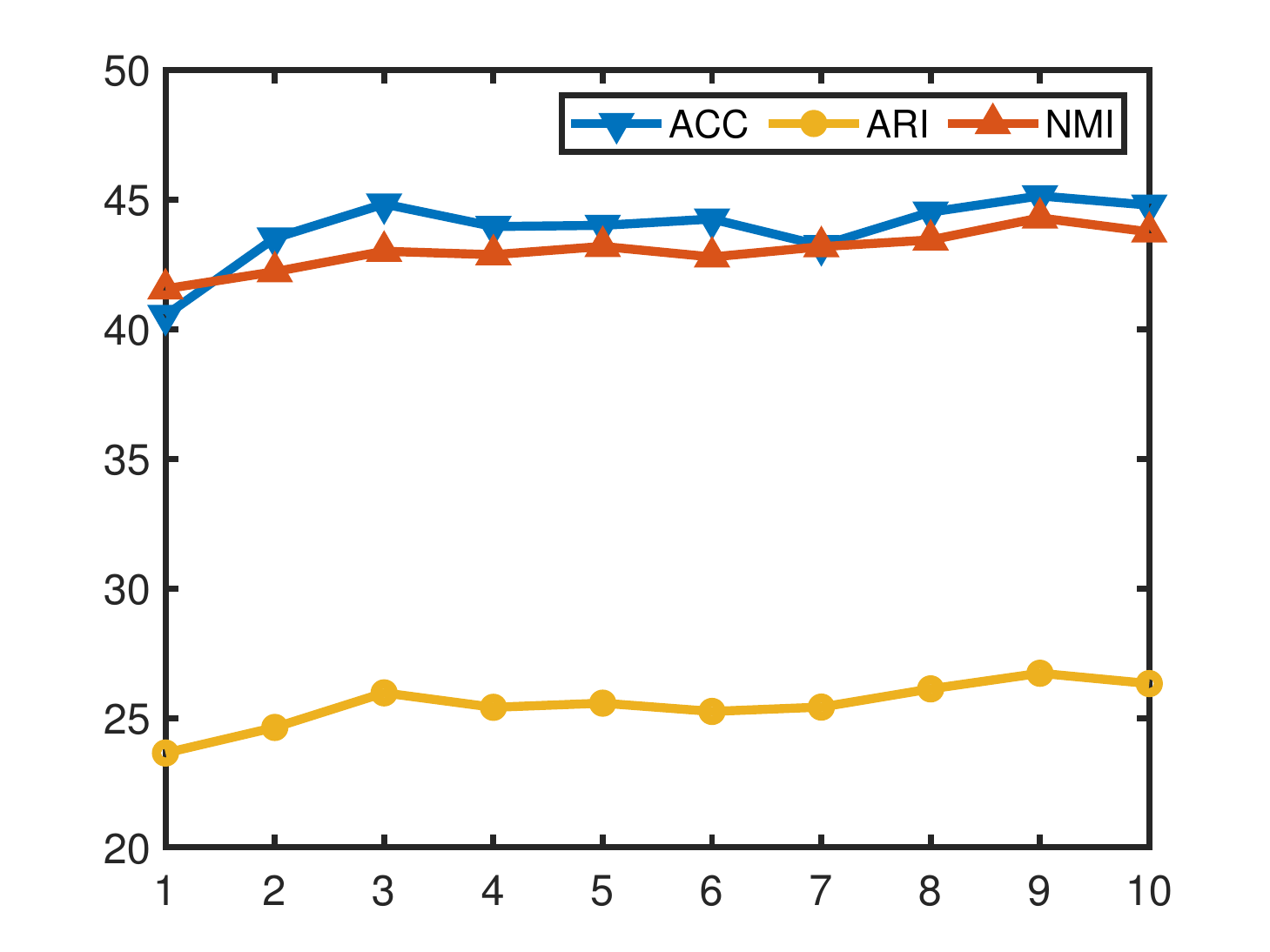}
}\hspace{-5mm}
\subfloat[$d_z$ on clustering task]
{
\label{subfig:dz_cluster}
\includegraphics[width=6cm]{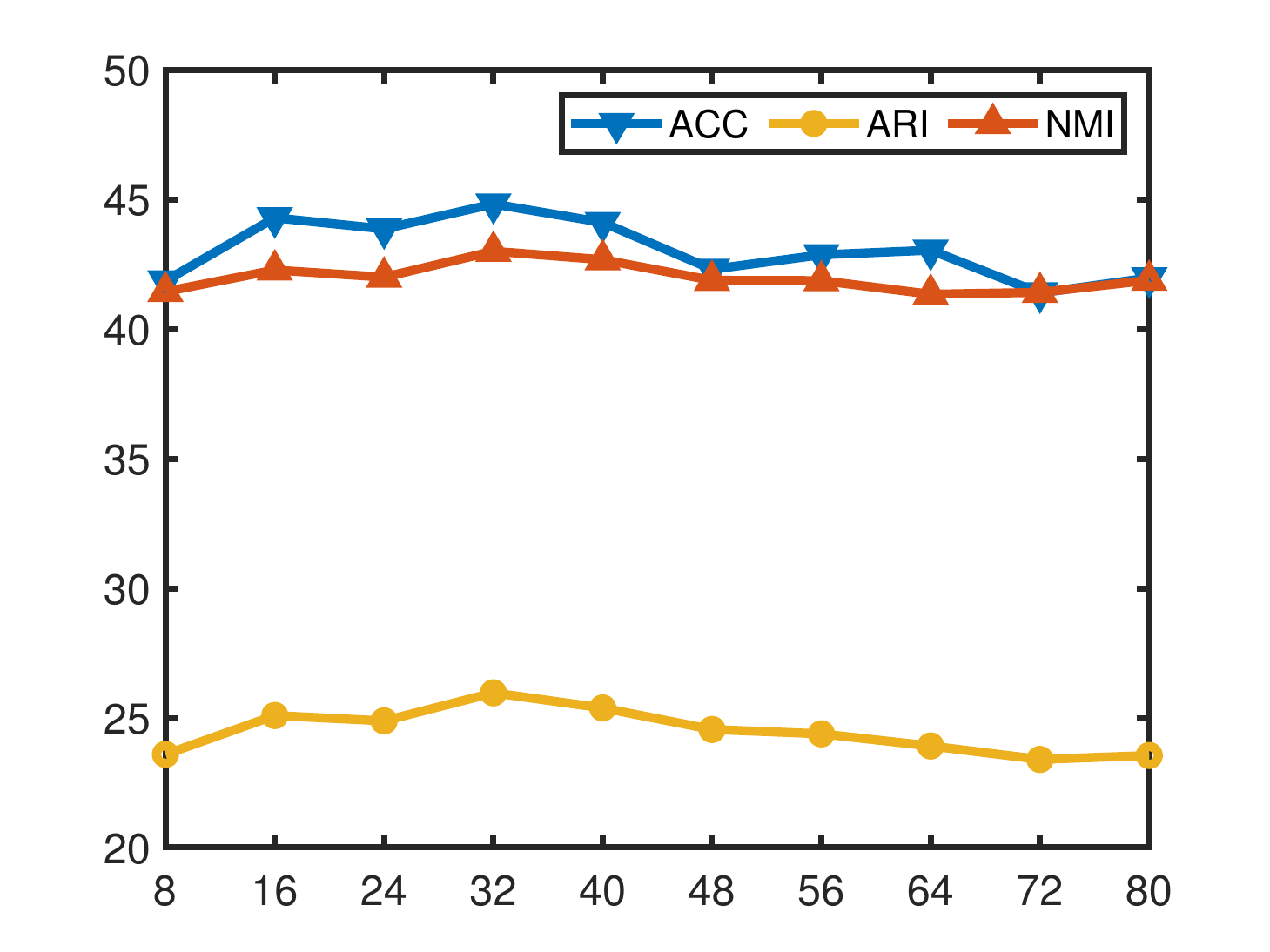}
}\hspace{-5mm}
\subfloat[$d_z$ on classification task]
{
\label{subfig:dz_classifer}
\includegraphics[width=6cm]{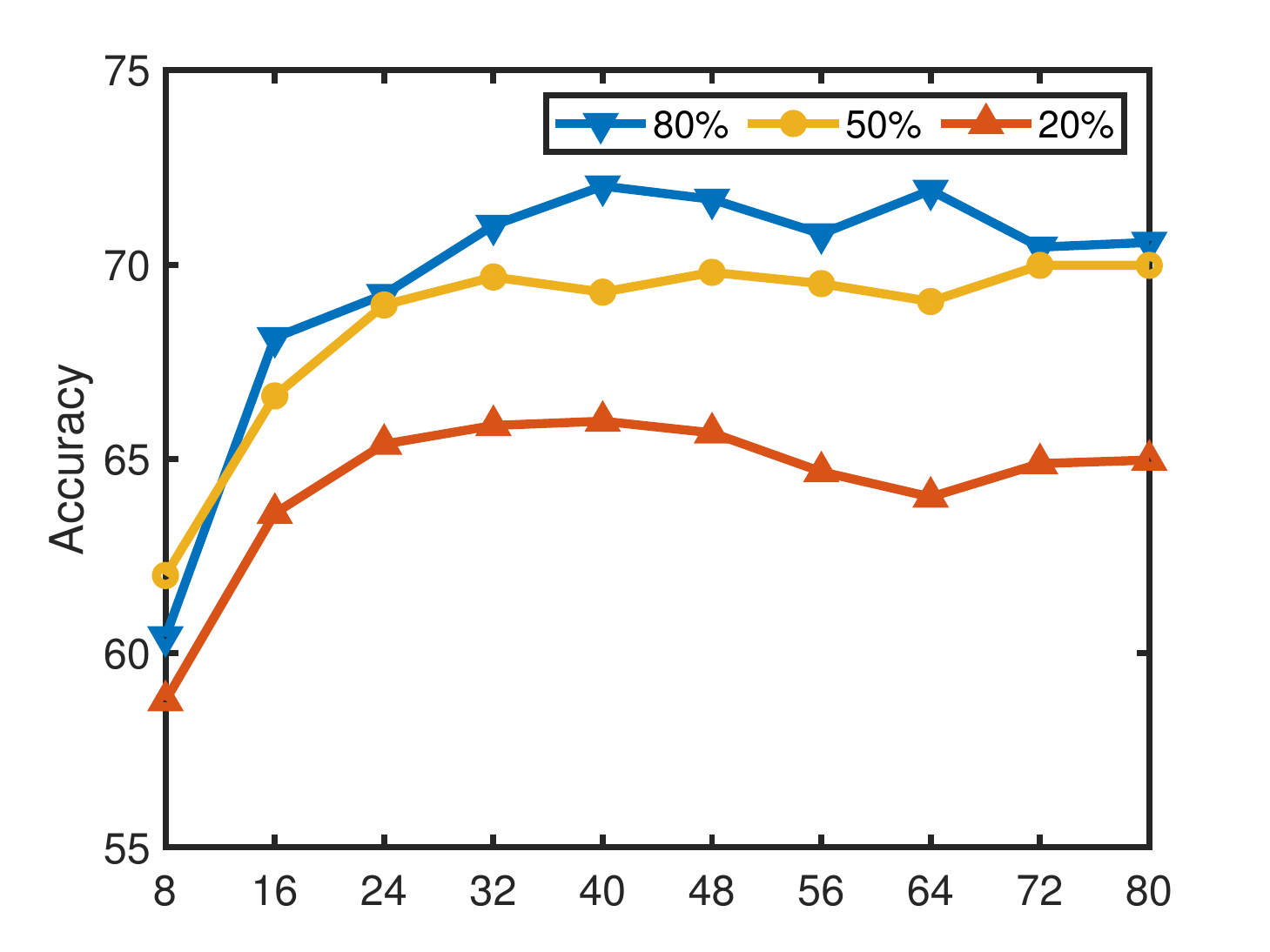}
}
\caption{Parameter analysis of $K$ and $d_z$ on Scene-15 dataset}
\label{fig:ParameterAnalyses}
\end{figure*} 

\begin{figure*}[t]
\centering
\subfloat[Scene-15]
{
\label{subfig:scene15_bs}
\includegraphics[width=9cm]{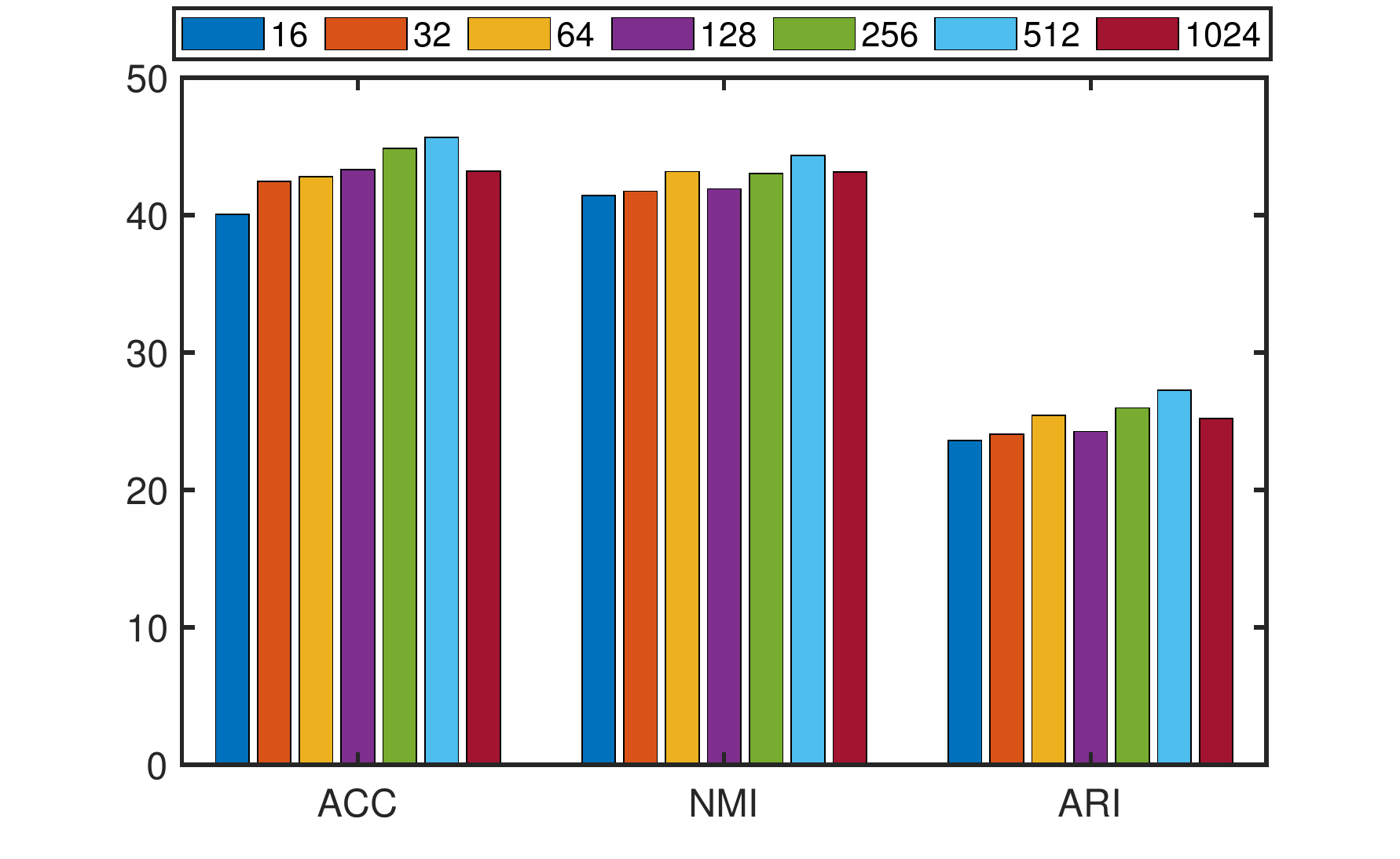}
}\hspace{-10mm}
\subfloat[100Leaves]
{
\label{subfig:100leaves_bs}
\includegraphics[width=9cm]{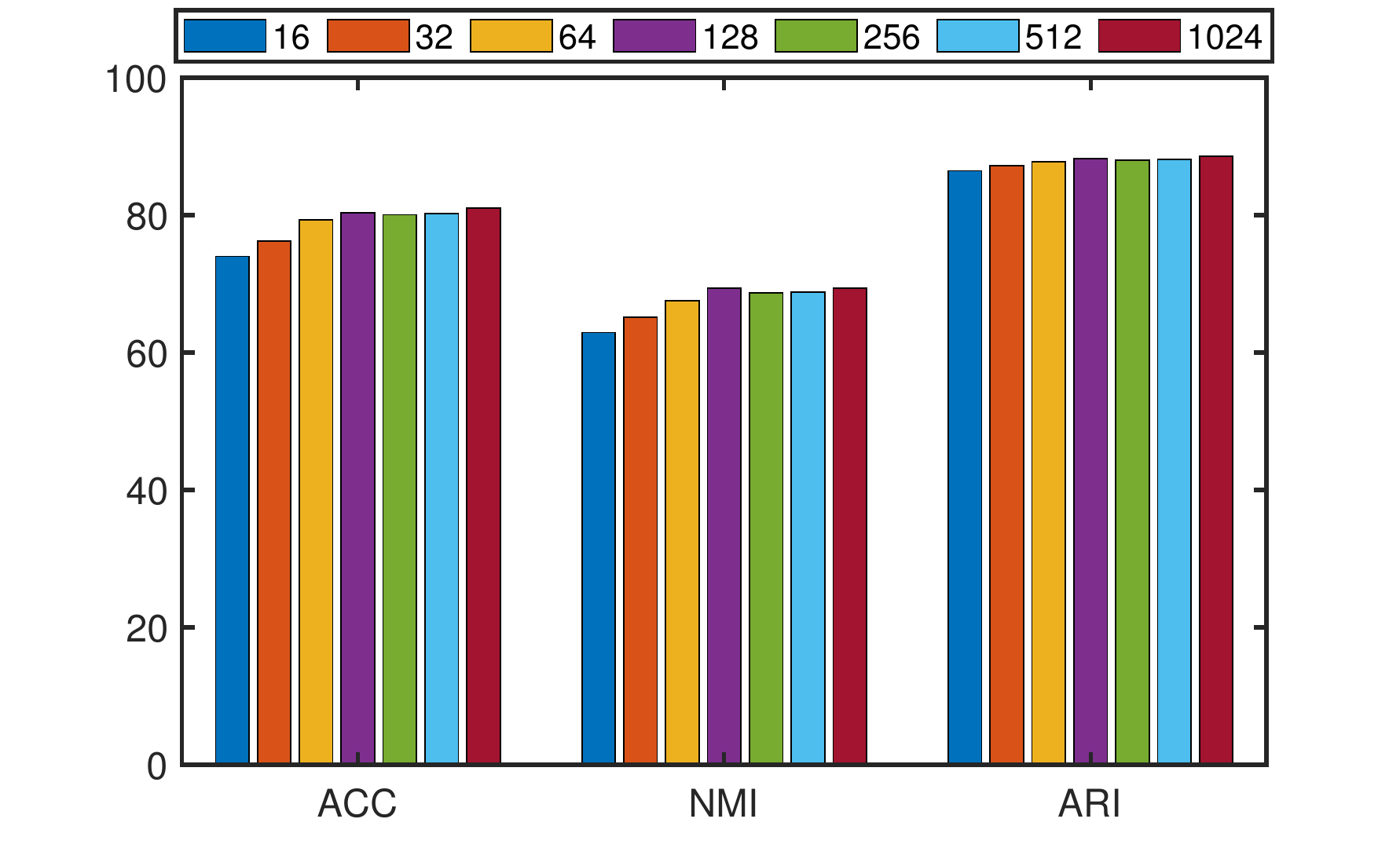}
}
\caption{Parameter analysis of $BatchSize$ on Scene-15 and 100Leaves datasets}
\label{fig:BATCHSIZEAnalyses}
\end{figure*} 

\subsection{Parameter Analyses}

\textit{\textbf{Loss weight parameter $\lambda$.}}
We vary the loss weight parameter $\lambda$ in Eq.~\ref{eq:all} from 1e-6 to 1000 to test its effect on clustering performance, and the results are shown in Fig.~\ref{fig:lambda}. From the results, we can notice that CIRCLE can obtain stable results when the $\lambda$ is within the range from 1e-3 to 2e-2, basically. When $\lambda$ is less than 1e-4, $L_{REC}$ is dominant, and the model gradually degenerates to use only $L_{REC}$. The clustering results drop significantly until close to the performance of using only $L_{REC}$ (see Table.~\ref{tab:ablation}). In contrast, a large $\lambda$ makes $L_{CGC}$ dominate the training phase. With the increase of the weight parameter, the performances start to drop from 5e-2 and gradually close to the performance of using only $L_{CGC}$ (see Table.~\ref{tab:ablation}).

\textit{\textbf{Number of aligned neighbors $K$.}}
The number of nearest neighbors is an important parameter in constructing KNN graphs and generally has an impact on the performance of graph-based representation learning methods. In order to check the impact of the number of the aligned nearest neighbors $K$, we study the performance of CIRCLE with various numbers of $K$ ranging from 1 to 10 in Fig.~\ref{subfig:k}. It can be seen from the figure that all metrics increase as $K$ increases from 1 to 3, and when $K \in [3, 10]$, the metrics are basically stable. It proves that our method is not sensitive to $K$ and can learn a stable representation even when there are fewer relational samples or some noisy edges in the graph. However, as $K$ increases, the number of positive, and negative pairs increases and the training time also grows. In our implementations, we set $K = 3$ for all the datasets.

\textit{\textbf{Dimension of the latent representations $d_z$.}}
We then investigate the clustering and classification performance of the learned representations with different dimensions of the latent representation $d_z$. From Fig. ~\ref{subfig:dz_cluster}-\ref{subfig:dz_classifer}, we have the following observations: (1) The clustering performance is relatively stable and starts to degrade slowly when the dimensionality is greater than 48. Empirically, higher dimensionality representations can contain more information that is suitable for complex downstream tasks. However, for the clustering task, the high-dimensional representations increase difficulties for subsequent conventional clustering methods. (2) The classification performance increases obviously first and then reaches a stable point when $d_z \ge 32$. For the classification task, too small dimensionality cannot contain enough information to train the classifier, which makes classification performance worse. When the dimension is too large, some redundant information is generated, which can not improve the performance of classification but can give a high computational complexity. For a fair comparison, we set $d_z = 32$ in all our experiments.

\textit{\textbf{BatchSize.}} 
In the training process of contrastive learning methods, the batch size typically has an impact on performance because the number of positive and negative pairs and the batch size are related. In general, as the batch size increases, the number of positive and negative pairs built becomes larger and performance improves. However, a larger batch size brings a larger memory usage. We vary the $BatchSize$ from 16 to 1024 to test its effect on clustering performance, and the results are shown in Fig.~\ref{fig:BATCHSIZEAnalyses}. From the results, we can notice that the clustering performance increases slowly with increasing batch size basically. For the Scene-15 dataset, the clustering performance improves with increasing batch size when the batch size is less than 512. For the BBCSport dataset, the clustering performance first increases as the batch size increases and then stabilizes after the batch size is greater than 128. Therefore, we choose the $BatchSize$ from 512 and 1024 for each dataset in our experiments.

\section{Conclusion}
\label{SECConclusion}
In this paper, a new cross-view graph contrastive representation learning framework is proposed to address problems in partially aligned multi-view learning. There are two main challenges in partially aligned multi-view representation learning. First, it is hard to find correspondence between heterogeneous multi-source data. Second, it is unstable to train a model to exploit multi-view information to learn discriminative without any prior. To solve the above two problems, we devise a cross-view graph contrastive learning (CGC) module to optimize the representations learned by view-specific autoencoders. By introducing the CGC module, CIRCLE can perform cluster-level alignment and representation learning in one stage. Additionally, the construction of cross-view relation graphs enables our model to be unconstrained by the number of data modalities/sources. We conduct extensive comparison experiments on clustering and classification tasks and ablation studies to validate the superiority of the model and the effectiveness of each component. In the future, we plan to further solve the noise problem in the relation graphs and optimize the fusion between multiple sources of information.

\ifCLASSOPTIONcompsoc
  \section*{Acknowledgments}
\else
  \section*{Acknowledgment}
\fi

This work was supported in part by the the National Key Research and Development of China (No. 2018AAA0102100), and the National Natural Science Foundation of China (No. U1936212).

\bibliographystyle{IEEEtran}
\bibliography{mybib}

\begin{thebibliography}{10}
\providecommand{\url}[1]{#1}
\csname url@samestyle\endcsname
\providecommand{\newblock}{\relax}
\providecommand{\bibinfo}[2]{#2}
\providecommand{\BIBentrySTDinterwordspacing}{\spaceskip=0pt\relax}
\providecommand{\BIBentryALTinterwordstretchfactor}{4}
\providecommand{\BIBentryALTinterwordspacing}{\spaceskip=\fontdimen2\font plus
\BIBentryALTinterwordstretchfactor\fontdimen3\font minus
  \fontdimen4\font\relax}
\providecommand{\BIBforeignlanguage}[2]{{%
\expandafter\ifx\csname l@#1\endcsname\relax
\typeout{** WARNING: IEEEtran.bst: No hyphenation pattern has been}%
\typeout{** loaded for the language `#1'. Using the pattern for}%
\typeout{** the default language instead.}%
\else
\language=\csname l@#1\endcsname
\fi
#2}}
\providecommand{\BIBdecl}{\relax}
\BIBdecl

\bibitem{conf/colt/BlumM98}
A.~Blum and T.~M. Mitchell, ``Combining labeled and unlabeled data with
  co-training,'' in \emph{Proceedings of the Eleventh Annual Conference on
  Computational Learning Theory}, 1998, pp. 92--100.

\bibitem{conf/cvpr/ZhangLF19}
C.~Zhang, Y.~Liu, and H.~Fu, ``Ae2-nets: Autoencoder in autoencoder networks,''
  in \emph{Proceedings of {IEEE} Conference on Computer Vision and Pattern
  Recognition}, 2019, pp. 2577--2585.

\bibitem{9258396}
C.~Zhang, Y.~Cui, Z.~Han, J.~T. Zhou, H.~Fu, and Q.~Hu, ``Deep partial
  multi-view learning,'' \emph{{IEEE} Trans. Pattern Anal. Mach. Intell.}, pp.
  1--1, 2020.

\bibitem{conf/ijcai/ZhangLLHLZ18}
C.~Zhang, Y.~Liu, Y.~Liu, Q.~Hu, X.~Liu, and P.~Zhu, ``{FISH-MML:} fisher-hsic
  multi-view metric learning,'' in \emph{Proceedings of the Twenty-Seventh
  International Joint Conference on Artificial Intelligence}, 2018, pp.
  3054--3060.

\bibitem{cca}
H.~Hotelling, ``Relations between two sets of variates,'' \emph{Biometrika},
  vol.~28, pp. 321--377, 1936.

\bibitem{conf/icml/dAspremontBG07}
A.~d'Aspremont, F.~R. Bach, and L.~E. Ghaoui, ``Full regularization path for
  sparse principal component analysis,'' in \emph{Proceedings of the
  Twenty-Fourth International Conference on Machine Learning}, ser. {ACM}
  International Conference Proceeding Series, vol. 227, 2007, pp. 177--184.

\bibitem{journals/ijns/LaiF00}
P.~L. Lai and C.~Fyfe, ``Kernel and nonlinear canonical correlation analysis,''
  \emph{Int. J. Neural Syst.}, vol.~10, no.~5, pp. 365--377, 2000.

\bibitem{conf/cvpr/YanM15}
F.~Yan and K.~Mikolajczyk, ``Deep correlation for matching images and text,''
  in \emph{Proceedings of {IEEE} Conference on Computer Vision and Pattern
  Recognition}, 2015, pp. 3441--3450.

\bibitem{conf/mm/0001ZZWFXZ20}
J.~Wen, Z.~Zhang, Z.~Zhang, Z.~Wu, L.~Fei, Y.~Xu, and B.~Zhang, ``Dimc-net:
  Deep incomplete multi-view clustering network,'' in \emph{Proceedings of the
  28th {ACM} International Conference on Multimedia}, 2020, pp. 3753--3761.

\bibitem{conf/cvpr/0001GLL0021}
Y.~Lin, Y.~Gou, Z.~Liu, B.~Li, J.~Lv, and X.~Peng, ``{COMPLETER:} incomplete
  multi-view clustering via contrastive prediction,'' in \emph{Proceedings of
  {IEEE} Conference on Computer Vision and Pattern Recognition}, 2021, pp.
  11\,174--11\,183.

\bibitem{conf/nips/Huang0ZL020}
Z.~Huang, P.~Hu, J.~T. Zhou, J.~Lv, and X.~Peng, ``Partially view-aligned
  clustering,'' in \emph{Proceedings of Advances in Neural Information
  Processing Systems}, 2020.

\bibitem{conf/cvpr/YangL0L0021}
M.~Yang, Y.~Li, Z.~Huang, Z.~Liu, P.~Hu, and X.~Peng, ``Partially view-aligned
  representation learning with noise-robust contrastive loss,'' in
  \emph{Proceedings of {IEEE} Conference on Computer Vision and Pattern
  Recognition}, 2021, pp. 1134--1143.

\bibitem{aaai/LiJZ14}
S.~Li, Y.~Jiang, and Z.~Zhou, ``Partial multi-view clustering,'' in
  \emph{Proceedings of the Thirty-Third {AAAI} Conference on Artificial
  Intelligence}, 2014, pp. 1968--1974.

\bibitem{conf/aaai/HuC19}
M.~Hu and S.~Chen, ``One-pass incomplete multi-view clustering,'' in
  \emph{Proceedings of the Thirty-Third {AAAI} Conference on Artificial
  Intelligence}, 2019, pp. 3838--3845.

\bibitem{journals/corr/GoodfellowPMXWOCB14}
I.~J. Goodfellow, J.~Pouget{-}Abadie, M.~Mirza, B.~Xu, D.~Warde{-}Farley,
  S.~Ozair, A.~C. Courville, and Y.~Bengio, ``Generative adversarial
  networks,'' \emph{[Online]. Available: arXiv:1406.2661}, 2014.

\bibitem{9462394}
X.~Liu, F.~Zhang, Z.~Hou, L.~Mian, Z.~Wang, J.~Zhang, and J.~Tang,
  ``Self-supervised learning: Generative or contrastive,'' \emph{IEEE Trans.
  Knowl. Data Eng.}, pp. 1--20, (Early Access) 2021.

\bibitem{conf/nips/Tian0PKSI20}
Y.~Tian, C.~Sun, B.~Poole, D.~Krishnan, C.~Schmid, and P.~Isola, ``What makes
  for good views for contrastive learning?'' in \emph{Proceedings of Advances
  in Neural Information Processing Systems}, 2020.

\bibitem{9667296}
Q.~Lin, J.~Liu, L.~Zhang, Y.~Pan, X.~Hu, F.~Xu, and H.~Zeng, ``Contrastive
  graph representations for logical formulas embedding,'' \emph{IEEE Trans.
  Knowl. Data Eng.}, pp. 1--12, (Early Access) 2021.

\bibitem{conf/cvpr/HadsellCL06}
R.~Hadsell, S.~Chopra, and Y.~LeCun, ``Dimensionality reduction by learning an
  invariant mapping,'' in \emph{Proceedings of {IEEE} Computer Society
  Conference on Computer Vision and Pattern Recognition}, 2006, pp. 1735--1742.

\bibitem{journals/jmlr/GutmannH10}
M.~Gutmann and A.~Hyv{\"{a}}rinen, ``Noise-contrastive estimation: {A} new
  estimation principle for unnormalized statistical models,'' in
  \emph{Proceedings of the Thirteenth International Conference on Artificial
  Intelligence and Statistics}, ser. {JMLR} Proceedings, vol.~9, 2010, pp.
  297--304.

\bibitem{conf/cvpr/He0WXG20}
K.~He, H.~Fan, Y.~Wu, S.~Xie, and R.~B. Girshick, ``Momentum contrast for
  unsupervised visual representation learning,'' in \emph{Proceedings of
  {IEEE/CVF} Conference on Computer Vision and Pattern Recognition}, 2020, pp.
  9726--9735.

\bibitem{conf/icml/ChenK0H20}
T.~Chen, S.~Kornblith, M.~Norouzi, and G.~E. Hinton, ``A simple framework for
  contrastive learning of visual representations,'' in \emph{Proceedings of the
  37th International Conference on Machine Learning}, ser. Proceedings of
  Machine Learning Research, vol. 119, 2020, pp. 1597--1607.

\bibitem{conf/fgr/0001TSS20}
V.~Sharma, M.~Tapaswi, M.~S. Sarfraz, and R.~Stiefelhagen, ``Clustering based
  contrastive learning for improving face representations,'' in
  \emph{Proceedings of 15th {IEEE} International Conference on Automatic Face
  and Gesture Recognition}, 2020, pp. 109--116.

\bibitem{conf/iccv/ZhongW0HDNL021}
H.~Zhong, J.~Wu, C.~Chen, J.~Huang, M.~Deng, L.~Nie, Z.~Lin, and X.~Hua,
  ``Graph contrastive clustering,'' in \emph{Proceedings of International
  Conference on Computer Vision}, 2021, pp. 9204--9213.

\bibitem{8471216}
Y.~Li, M.~Yang, and Z.~Zhang, ``A survey of multi-view representation
  learning,'' \emph{IEEE Trans. Knowl. Data Eng.}, vol.~31, no.~10, pp.
  1863--1883, 2019.

\bibitem{8493362}
X.~Fu, K.~Huang, E.~E. Papalexakis, H.~A. Song, P.~Talukdar, N.~D.
  Sidiropoulos, C.~Faloutsos, and T.~Mitchell, ``Efficient and distributed
  generalized canonical correlation analysis for big multiview data,''
  \emph{IEEE Trans. Knowl. Data Eng.}, vol.~31, no.~12, pp. 2304--2318, 2019.

\bibitem{7123622}
Y.~Luo, D.~Tao, K.~Ramamohanarao, C.~Xu, and Y.~Wen, ``Tensor canonical
  correlation analysis for multi-view dimension reduction,'' \emph{IEEE Trans.
  Knowl. Data Eng.}, vol.~27, no.~11, pp. 3111--3124, 2015.

\bibitem{conf/icml/AndrewABL13}
G.~Andrew, R.~Arora, J.~A. Bilmes, and K.~Livescu, ``Deep canonical correlation
  analysis,'' in \emph{Proceedings of the 30th International Conference on
  Machine Learning}, ser. {JMLR} Workshop and Conference Proceedings, vol.~28,
  2013, pp. 1247--1255.

\bibitem{conf/icml/WangALB15}
W.~Wang, R.~Arora, K.~Livescu, and J.~A. Bilmes, ``On deep multi-view
  representation learning,'' in \emph{Proceedings of the 32nd International
  Conference on Machine Learning}, ser. {JMLR} Workshop and Conference
  Proceedings, vol.~37, 2015, pp. 1083--1092.

\bibitem{conf/ijcai/WenZ0ZFX20}
J.~Wen, Z.~Zhang, Y.~Xu, B.~Zhang, L.~Fei, and G.~Xie, ``Cdimc-net: Cognitive
  deep incomplete multi-view clustering network,'' in \emph{Proceedings of the
  Twenty-Ninth International Joint Conference on Artificial Intelligence},
  C.~Bessiere, Ed., 2020, pp. 3230--3236.

\bibitem{journals/jmlr/PedregosaVGMTGBPWDVPCBPD11}
F.~Pedregosa, G.~Varoquaux, A.~Gramfort, V.~Michel, B.~Thirion, O.~Grisel,
  M.~Blondel, P.~Prettenhofer, R.~Weiss, V.~Dubourg, J.~VanderPlas, A.~Passos,
  D.~Cournapeau, M.~Brucher, M.~Perrot, and E.~Duchesnay, ``Scikit-learn:
  Machine learning in python,'' \emph{J. Mach. Learn. Res.}, vol.~12, pp.
  2825--2830, 2011.

\bibitem{conf/cvpr/LiPT05}
L.~Fei{-}Fei and P.~Perona, ``A bayesian hierarchical model for learning
  natural scene categories,'' in \emph{Proceedings of {IEEE} Computer Society
  Conference on Computer Vision and Pattern Recognition}, 2005, pp. 524--531.

\bibitem{journals/pami/ZhangLSSS19}
Z.~Zhang, L.~Liu, F.~Shen, H.~T. Shen, and L.~Shao, ``Binary multi-view
  clustering,'' \emph{{IEEE} Trans. Pattern Anal. Mach. Intell.}, vol.~41,
  no.~7, pp. 1774--1782, 2019.

\bibitem{conf/nips/AminiUG09}
M.~Amini, N.~Usunier, and C.~Goutte, ``Learning from multiple partially
  observed views - an application to multilingual text categorization,'' in
  \emph{Proceedings of Advances in Neural Information Processing Systems},
  2009, pp. 28--36.

\bibitem{conf/kdd/YuTWG21}
H.~Yu, J.~Tang, G.~Wang, and X.~Gao, ``A novel multi-view clustering method for
  unknown mapping relationships between cross-view samples,'' in
  \emph{Proceedings of the 27th {ACM} {SIGKDD} Conference on Knowledge
  Discovery and Data Mining}, 2021, pp. 2075--2083.

\bibitem{conf/ijcai/WangZLYZ19}
H.~Wang, L.~Zong, B.~Liu, Y.~Yang, and W.~Zhou, ``Spectral perturbation meets
  incomplete multi-view data,'' in \emph{Proceedings of the Twenty-Eighth
  International Joint Conference on Artificial Intelligence}, 2019, pp.
  3677--3683.

\bibitem{journals/tcyb/ZhanZGW18}
K.~Zhan, C.~Zhang, J.~Guan, and J.~Wang, ``Graph learning for multiview
  clustering,'' \emph{{IEEE} Trans. Cybern.}, vol.~48, no.~10, pp. 2887--2895,
  2018.

\bibitem{journals/tip/ZhanNWY19}
K.~Zhan, F.~Nie, J.~Wang, and Y.~Yang, ``Multiview consensus graph
  clustering,'' \emph{{IEEE} Trans. Image Process.}, vol.~28, no.~3, pp.
  1261--1270, 2019.

\bibitem{journals/tkde/WangYL20}
H.~Wang, Y.~Yang, and B.~Liu, ``{GMC:} graph-based multi-view clustering,''
  \emph{IEEE Trans. Knowl. Data Eng.}, vol.~32, no.~6, pp. 1116--1129, 2020.

\bibitem{conf/aaai/KangZZSHX20}
Z.~Kang, W.~Zhou, Z.~Zhao, J.~Shao, M.~Han, and Z.~Xu, ``Large-scale multi-view
  subspace clustering in linear time,'' in \emph{Proceedings of the
  Thirty-Fourth {AAAI} Conference on Artificial Intelligence}, 2020, pp.
  4412--4419.

\bibitem{/conf/iccv/opmc}
J.~Liu, X.~Liu, Y.~Yang, L.~Liu, S.~Wang, W.~Liang, and J.~Shi, ``One-pass
  multi-view clustering for large-scale data,'' in \emph{Proceedings of
  IEEE/CVF International Conference on Computer Vision}, 2021, pp.
  12\,324--12\,333.

\bibitem{conf/mm/SunZWZTLZW21}
M.~Sun, P.~Zhang, S.~Wang, S.~Zhou, W.~Tu, X.~Liu, E.~Zhu, and C.~Wang,
  ``Scalable multi-view subspace clustering with unified anchors,'' in
  \emph{Proceedings of {ACM} Multimedia Conference}, 2021, pp. 3528--3536.

\bibitem{conf/soda/ArthurV07}
D.~Arthur and S.~Vassilvitskii, ``k-means++: the advantages of careful
  seeding,'' in \emph{Proceedings of the Eighteenth Annual {ACM-SIAM} Symposium
  on Discrete Algorithms}, N.~Bansal, K.~Pruhs, and C.~Stein, Eds., 2007, pp.
  1027--1035.

\bibitem{Platt99probabilisticoutputs}
J.~C. Platt, ``Probabilistic outputs for support vector machines and
  comparisons to regularized likelihood methods,'' in \emph{ADVANCES IN LARGE
  MARGIN CLASSIFIERS}, 1999, pp. 61--74.

\bibitem{jmlr/laurens08}
L.~V.~D. Maaten and G.~Hinton, ``Visualizing data using t-sne,'' \emph{J. Mach.
  Learn. Res.}, vol.~9, no. 2605, pp. 2579--2605, 2008.

\end{thebibliography}

\end{document}